\documentclass[a4paper,runningheads]{llncs}

\usepackage{alltt}
\usepackage{amssymb}
\usepackage{caption}
\usepackage{paralist}
\usepackage{nicefrac}
\usepackage{booktabs}
\usepackage{xspace}
\usepackage{amsmath}
\usepackage{marvosym}
\usepackage{wasysym}
\usepackage{verbatim}
\usepackage{float}
\usepackage{hyperref}
\usepackage{multirow}
\usepackage{cite}
\usepackage[capitalize]{cleveref}
\usepackage[per-mode=symbol,detect-all]{siunitx}
\usepackage{graphics}
\usepackage{amssymb}
\usepackage{enumitem}
\usepackage[table]{xcolor}
\usepackage{booktabs}
\usepackage{colortbl}
\usepackage{ifthen}
\usepackage{mathdots}
\usepackage{arydshln}
\usepackage[outline]{contour}
\usepackage{graphicx}
\usepackage{wrapfig}
\usepackage{subfig}

\definecolor{navy}{RGB}{0,0,128}

\usepackage{tikz,ifthen,pgfplots}
\usetikzlibrary{arrows,trees,backgrounds,automata,shapes,decorations,plotmarks,fit,calc,positioning,shadows,chains}
\usetikzlibrary{quotes,arrows.meta}
\tikzstyle{every pin edge}=[<-,shorten <=1pt]
\tikzstyle{neuron}=[circle,fill=black!25,minimum size=17pt,inner sep=0pt]
\tikzstyle{input neuron}=[neuron, fill=green!50]
\tikzstyle{output neuron}=[neuron, fill=red!50]
\tikzstyle{hidden neuron}=[neuron, fill=blue!50]
\tikzstyle{small neuron}        =[hidden neuron, draw, minimum size=15pt]
\tikzstyle{small input neuron}  =[input neuron , draw, minimum size=15pt]
\tikzstyle{small output neuron} =[output neuron, draw, minimum size=15pt]
\tikzstyle{annot} = [text width=4em, text centered]
\tikzstyle{nnedge} = [-{stealth},shorten >=0.1cm, shorten <=0.05cm,line width=0.8pt,black]
\tikzstyle{edge} = [->,line width = 0.3pt, shorten >=0.2cm]
\tikzstyle{edgeWide} = [->,line width = 2pt, , shorten >=0.2cm]
\usetikzlibrary{calc}

\tikzset{every picture/.style={line width=0.75pt}} 
\tikzstyle{BadSquare}=[rectangle,fill=red!30!white,minimum size=25pt,inner 
sep=0pt]
\tikzstyle{InitSquare}=[rectangle,fill=green!30!white,minimum size=25pt,inner 
sep=0pt]

\newcommand{\mysubsection}[1]{\medskip\noindent\textbf{#1}}

\newcommand{\relu}{\text{ReLU}\xspace}

\newcommand{\sat}{\texttt{SAT}\xspace}
\newcommand{\unsat}{\texttt{UNSAT}\xspace}
\newcommand{\timeout}{\texttt{TIMEOUT}\xspace}
\newcommand{\fail}{\texttt{FAIL}\xspace}

\newcommand{\forwardOutput}{\texttt{FORWARD}\xspace}
\newcommand{\leftOutput}{\texttt{LEFT}\xspace}
\newcommand{\rightOutput}{\texttt{RIGHT}\xspace}

\newcommand{\forwardCollision}{\texttt{FORWARD COLLISION}\xspace}
\newcommand{\leftCollision}{\texttt{LEFT COLLISION}\xspace}
\newcommand{\rightCollision}{\texttt{RIGHT COLLISION}\xspace}

\newcommand{\alternatingLoop}{\texttt{ALTERNATING LOOP}\xspace}
\newcommand{\leftCycle}{\texttt{LEFT CYCLE}\xspace}
\newcommand{\rightCycle}{\texttt{RIGHT CYCLE}\xspace}

\newcommand{\marabou}{\textit{Marabou}\xspace}

\newcommand{\memout}{\texttt{MEMOUT}\xspace}

\usepackage{bussproofs}
\usepackage{gensymb}

\newif\ifcomments
\commentstrue

\newif\ifoutline
\outlinefalse

\newif\iflong
\longtrue

\renewcommand{\paragraph}[1]{\vspace{1mm}\noindent{\bf #1}\ }

\usepackage{lipsum}

\authorrunning{G. Amir, D. Corsi et al.}

\begin{document}
	
	\title{Verifying Learning-Based\\Robotic Navigation Systems}
	
	\author{
		Guy Amir\inst{1,*} \and
		Davide Corsi\inst{2,*} \and
		Raz Yerushalmi\inst{1,3} \and
		Luca Marzari\inst{2} \and \\
		David Harel\inst{3} \and
		Alessandro Farinelli\inst{2} \and
		Guy Katz\inst{1}		
	}

	\institute{
		The Hebrew University of Jerusalem, Jerusalem, Israel \\
		\email{ \{guyam, guykatz\}@cs.huji.ac.il} \\
		\and
		University of Verona, Verona, Italy \\
		\email{ \{davide.corsi, luca.marzari, alessandro.farinelli\}@univr.it}\\
		\and
		The Weizmann Institute of Science, Rehovot, Israel \\
		\email{ \{raz.yerushalmi, david.harel\}@weizmann.ac.il}\\
	}

	\maketitle

	\let\svthefootnote\thefootnote
	\let\thefootnote\relax\footnotetext{[*] Both authors contributed equally.}
	\let\thefootnote\svthefootnote
	\addtocounter{footnote}{-3}
	
	\begin{abstract} 		
		Deep reinforcement learning (DRL) has become a dominant
		deep-learning paradigm for tasks where complex policies
		are learned within reactive systems. 
                Unfortunately, these policies are known to be
                susceptible to bugs. 
                Despite significant progress in DNN 
		verification, there has been little work demonstrating the use of 
		modern verification tools on real-world, DRL-controlled systems.
		In this case study, we attempt to begin bridging
                this gap, and focus on the important task of mapless
                robotic navigation --- a classic robotics problem, in
                which a robot, usually controlled by a DRL agent,
                needs to efficiently and safely navigate through an
                unknown arena towards a target. We demonstrate
                how modern verification engines can be used for
                effective \emph{model selection}, i.e., selecting the best 
                available policy for the robot in
                question from a pool of candidate policies.
                Specifically, we use verification to detect and rule
                out policies that may demonstrate suboptimal behavior,
                such as collisions and infinite loops.  We also apply
                verification to identify models with overly
                conservative behavior, thus allowing users to choose
                superior policies, which might be better at finding shorter
                paths to a target. To validate our work, we conducted
                extensive experiments on an actual robot, and
                confirmed that the suboptimal policies detected by our
                method were indeed flawed.  We also demonstrate the
                superiority of our verification-driven approach over
                state-of-the-art, gradient attacks.  Our work is the
                first to establish the usefulness of DNN
                verification in identifying and filtering out
                suboptimal DRL policies in real-world robots, and we
                believe that the methods presented here are applicable
                to a wide range of systems that incorporate
                deep-learning-based agents.
		
	\end{abstract}

	\section{Introduction}
	\label{sec:Introduction}

	In recent years, \emph{deep neural networks} (DNN) have become
	extremely popular, due to achieving state-of-the-art results in a
	variety of fields --- such as natural language
	processing~\cite{DeLi18}, image recognition~\cite{SiZi14}, autonomous
	driving~\cite{BoDeDwFiFlGoJaMoMuZhZhZhZi16}, and more. The immense
	success of these DNN models is owed in part to their ability to train on a
	fixed set of training samples drawn from some distribution, and then 
	\emph{generalize}, i.e., correctly handle inputs that they had not
	encountered previously.  Notably, \emph{deep reinforcement learning}
	(DRL)~\cite{Li17} has recently become a dominant paradigm for training
	DNNs that implement control policies for complex systems that operate
	within rich environments. One domain in which DRL controllers
	have been especially successful is robotics, and specifically ---
	robotic navigation, i.e., the complex task of efficiently
	navigating a robot through an arena, in order to safely reach a
	target~\cite{WaToFi19, ZhSpJo17}.
	
	Unfortunately, despite the immense success of DNNs, they have been
	shown to suffer from various safety 
	issues~\cite{KaBaDiJuKo17, 
		SzZaSuBrErGoFe13}. 
	For example, small
	perturbations to their inputs, which are either intentional or the
	result of noise, may cause 
	DNNs to react in unexpected ways~\cite{MoSeFaPa17}. 
	These inherent weaknesses, and
	others, are observed in almost every kind of neural network, and
	indicate a need for techniques that can supply formal guarantees
	regarding the safety 
	of the DNN in question.
	These weaknesses have also been observed in DRL
	systems~\cite{ElKaKaSc21,KaBaKaSc19,AmScKa21}, showing that even
	state-of-the-art DRL models may err miserably.
	
	To mitigate such safety issues, the verification
	community has recently developed a plethora of techniques and
	tools~\cite{KaBaDiJuKo17, GeMiDrTsChVe18, WaPeWhYaJa18,
		LyKoKoWoLiDa20, HuKwWaWu17, ZhShGuGuLeNa20, KoLoJaBl20,
		BaShShMeSa19, LoMa17, AvBlChHeKoPr19, IvCaWeAlPaLe20, DuJhSaTi18} for 
		formally 
		verifying that a DNN 
		model
	is
	safe to deploy. Given a DNN, these methods usually check whether the DNN: (i)
	behaves according to a prescribed requirement for \textit{all}
	possible inputs of interest; or (ii) violates the requirement, in which 
	case the
	verification tool also provides a counterexample.
	
	To date, despite the abundance of both DRL systems and DNN
        verification techniques, little work has been published on
        demonstrating the applicability and usefulness of verification
        techniques to real-world DRL systems. In this case study, we
        showcase the capabilities of DNN verification tools for
        analyzing DRL-based systems in the robotics domain --- 
        specifically, robotic navigation systems. To the best of our
        knowledge, this is the first attempt to demonstrate how
        off-the-shelf verification engines can be used to identify both
        \emph{unsafe} and \emph{suboptimal} DRL robotic controllers,
        that cannot be detected otherwise using existing, incomplete
        methods.  Our approach leverages existing DNN verifiers that
        can reason about single and multiple invocations of DRL
        controllers, and this allows us to conduct a
        verification-based model selection process --- through which
        we filter out models that could render the system unsafe.
	
	In addition to model 
	selection, 
	we demonstrate how verification methods allow gaining better
        insights into the DRL training process, by comparing the
        outcomes of different training methods and assessing how the
        models improve over additional training iterations.  We also
        compare our approach to gradient-based methods, and
        demonstrate the advantages of verification-based tools in this
        setting.
	We regard this as another step
	towards increasing the reliability and safety of DRL systems,
	which is one of the key challenges in modern machine learning~\cite{Gu17}; 
	and also as a step toward a more wholesome integration
	of verification techniques into the DRL development cycle.
	
	In order to validate our experiments, we
	conducted an extensive evaluation on a real-world, physical robot. Our
	results demonstrate that policies classified as suboptimal
	by our approach indeed exhibited unwanted behavior. This evaluation
	highlights the practical nature of our work; and is summarized in a
	short video clip~\cite{RobotYoutubeVideo}, which we strongly 
	encourage
	the reader to watch. In addition, our code and benchmarks are 
	available online~\cite{ArtifactRepository}.
	
	The rest of the paper is organized as follows.
	Section~\ref{sec:Background} contains background on DNNs, DRLs, and
	robotic controlling systems. In Section~\ref{sec:CaseStudy} we present
	our DRL robotic controller case study, and then elaborate on the
	various properties that we considered in
	Section~\ref{sec:PropertiesAndPolicySelection}. In
	Section~\ref{sec:Experiments} we present our experimental results, and
	use them to compare our approach with competing methods. Related work
	appears in Section~\ref{sec:RelatedWork}, and we conclude in
	Section~\ref{sec:Conclusion}.

	\section{Background}
	\label{sec:Background}

	\mysubsection{Deep Neural Networks.}  Deep neural networks
	(DNNs)~\cite{GoBeCo16} are computational, directed, graphs consisting
	of multiple layers.  By assigning values to the first layer of the
	graph and propagating them through the subsequent layers, the network
	computes either a label prediction (for a classification DNN) or a
	value (for a regression DNN), which is returned to the user. The values
	computed in each layer depend on values computed in previous
	layers, and also on the current layer's \textit{type}. Common layer
	types include the \emph{weighted sum} layer, in which each neuron is
	an affine transformation of the neurons from the preceding layer; as
	well as the popular \emph{rectified linear unit} (\emph{ReLU}) layer,
	where each node $y$ computes the value $y=\relu{}(x)=\max(0,x)$, based
	on a single node $x$ from the preceding layer to which it is
	connected.  The DRL systems that are the subject matter of this
	case study consist solely of weighted sum and \relu{} layers, although
	the techniques mentioned are suitable for DNNs with additional
	layer types, as we discuss later.

	\begin{wrapfigure}{r}{0.5\textwidth}
	\vspace{-1.5cm}
	\begin{center}
		\scalebox{0.75} {
			\def\layersep{2.0cm}
			\begin{tikzpicture}[shorten >=1pt,->,draw=black!50, node 
				distance=\layersep,font=\footnotesize]
				
				\node[input neuron] (I-1) at (0,-1) {$v^1_1$};
				\node[input neuron] (I-2) at (0,-2.5) {$v^2_1$};
				
				\node[hidden neuron] (H-1) at (\layersep,-1) {$v^1_2$};
				\node[hidden neuron] (H-2) at (\layersep,-2.5) {$v^2_2$};
				
				\node[hidden neuron] (H-3) at (2*\layersep,-1) {$v^1_3$};
				\node[hidden neuron] (H-4) at (2*\layersep,-2.5) {$v^2_3$};
				
				\node[output neuron] at (3*\layersep, -1.75) (O-1) 
				{$v^1_4$};
				
				\draw[nnedge] (I-1) --node[above] {$2$} (H-1);
				\draw[nnedge] (I-1) --node[above, pos=0.3] {$\ -4$} (H-2);
				\draw[nnedge] (I-2) --node[below, pos=0.3] {$5$} (H-1);
				\draw[nnedge] (I-2) --node[below] {$1$} (H-2);
				
				\draw[nnedge] (H-1) --node[above] {$\relu$} (H-3);
				\draw[nnedge] (H-2) --node[below] {$\relu$} (H-4);
				
				\draw[nnedge] (H-3) --node[above] {$2$} (O-1);
				\draw[nnedge] (H-4) --node[below] {$-1$} (O-1);

				\node[below=0.05cm of H-1] (b1) {$+1$};
				\node[below=0.05cm of H-2] (b2) {$-2$};
				
				\node[annot,above of=H-1, node distance=0.8cm] (hl1) 
				{Weighted 
					sum};
				\node[annot,above of=H-3, node distance=0.8cm] (hl2) {ReLU 
				};
				\node[annot,left of=hl1] {Input };
				\node[annot,right of=hl2] {Output };
			\end{tikzpicture}
		}	
	\end{center}
	\vspace{-4mm}
	\caption{A toy DNN.}
	\vspace{-6mm}
	\label{fig:toyDnn}
\end{wrapfigure}
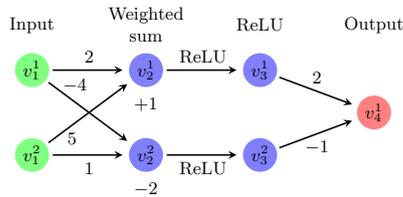
Fig.~\ref{fig:toyDnn} depicts a small example of a DNN. For input
	$V_1=[2, 3]^T$, the second (weighted sum) layer computes the values
	$V_2=[20,-7]^T$. In the third layer, the \relu{} functions are applied, 
	and the result is $V_3=[20,0]^T$. Finally, the network's single output
	is computed as a weighted sum: $V_4=[40]$.

	\mysubsection{Deep Reinforcement Learning.}  
	Deep reinforcement learning (DRL)~\cite{Li17} is a particular paradigm and 
	setting for training DNNs. In DRL, an \textit{agent} is trained to learn a 
	\emph{policy} $\pi$, which maps each possible \emph{environment state} $s$ 
	(i.e., the current observation of the agent) to an \emph{action} $a$. The 
	policy can have different interpretations among various learning 
	algorithms. For example, in some cases, $\pi$ represents a probability 
	distribution over the action space, while in others it encodes a function 
	that estimates a \textit{desirability score} over all the future actions 
	from 
	a state $s$.
	
	During training, at each discrete time-step $t\in\{0,1,2,\ldots\}$, a 
	\textit{reward} $r_t$ is presented to the agent, based on the action
	$a_t$ it performed at time-step $t$. Different DRL training algorithms
	leverage the reward in different ways, in order to optimize the DNN-agent's
	parameters during training. The general DNN architecture described above
	also characterizes DRL-trained DNNs; the uniqueness of the DRL paradigm
	lies in the training process, which is aimed at generating a DNN that
	computes a mapping $\pi$ that maximizes the \textit{expected
		cumulative discounted reward}
	$R_t=\mathbb{E}\big[\sum_{t}\gamma^{t}\cdot r_t\big]$. The
	\textit{discount factor}, $\gamma \in \big[0,1\big]$, is 
	a hyperparameter that controls the influence that past
	decisions have on the total expected reward.

	DRL training algorithms are typically divided into three
	categories~\cite{SuBa18}:
	
	\begin{enumerate}
		\item \textbf{Value-Based Algorithms.} 
		These algorithms attempt to learn a value function (called the 
		\emph{Q-function}) that 
		assigns a value 
		to  each $\langle$state,action$\rangle$ pair. 
		This iterative process relies on the \textit{Bellman 
		equation}~\cite{MnKaSi13} 
		to 
		update the function:
		$\mathbb{Q}^{\pi}(s_t, a_t) = r + \gamma 
		\max_{a_{t+1}}\mathbb{Q}^{\pi}(s_{t+1}, 
		a_{t+1})$. \textit{Double Deep Q-Network} (DDQN) is an optimized 
		implementation of this 
		algorithm~\cite{vHaGuSi16}.

		\item \textbf{Policy-Gradient Algorithms.} 
		This class contains algorithms that attempt to directly learn the 
		optimal 
		policy, instead of assessing the value function. The algorithms in this 
		class are typically based on the \textit{policy gradient 
			theorem}~\cite{SuMcSi99}. A	common implementation is the 
			\textit{Reinforce} 
		algorithm~\cite{ZhJoBr20}, 
		which aims to directly optimize the following objective function, over the 
		parameters $\theta$ of the DNN, through a gradient ascent process:
		$\nabla_\theta \mathbb{J(\pi_\theta)} = 
		\mathbb{E}[\sum_t^T\nabla_\theta \log{\pi_\theta(a_t|s_t)} \cdot r_t].$
		For additional details, see~\cite{ZhJoBr20}.
		
		\item
		\textbf{Actor-Critic Algorithms.} 
		This family of hybrid algorithms combines the two previous approaches. 
		The 
		key idea is to use two different neural networks: a \textit{critic},
		which 
		learns the value function from the data, and an
                \textit{actor}, which
		iteratively improves the policy by maximizing the value function 
		learned by 
		the critic. A state-of-the-art implementation of this approach is 
		the 
		\textit{Proximal Policy Optimization} (PPO) algorithm~\cite{ShWoDh17}. 
		
	\end{enumerate}

	All of these approaches are commonly used in modern DRL; and each
	has its advantages and disadvantages.  For example, the
	value-based methods typically require only small sets of examples to
	learn from, but are unable to learn policies for continuous spaces of
	$\langle$state,action$\rangle$ pairs.  In contrast, the
	policy-gradient methods can learn continuous policies, but suffer from
	a low sample efficiency and large memory requirements. Actor-Critic 
	algorithms
	attempt to combine the benefits of value-based and
	policy-gradient methods, but suffer from high instability,
	particularly in the early stages of training, when the value function
	learned by the critic is unreliable.

	\mysubsection{DNN Verification and DRL Verification.}  A DNN
	verification algorithm receives as input~\cite{KaBaDiJuKo17}:
	\begin{inparaenum}[(i)]
		\item a trained DNN $N$;
		\item a precondition $P$ on the DNN's inputs, which limits their
		possible assignments to inputs of interest; and	
		\item a postcondition $Q$ on $N$'s output, which usually encodes the 
		\textit{negation} of the behavior we would like $N$ to exhibit on
		inputs that satisfy $P$.
	\end{inparaenum}
	The verification algorithm then searches for a concrete input $x_0$ that
	satisfies $P(x_0) \wedge Q(N(x_0))$, and returns one of the following
	outputs:
	\begin{inparaenum}[(i)]
		\item \sat, along with a concrete input $x_0$ that satisfies the given
		constraints; or
		\item \unsat, indicating that no such $x_0$ exists.
	\end{inparaenum}
	When $Q$ encodes the negation of the required property, a \sat result
	indicates that the property is violated (and the returned input $x_0$
	triggers a bug), while an \unsat result indicates that the property
	holds.

	For example, suppose we wish to verify that the DNN in
	Fig.~\ref{fig:toyDnn} always outputs a value strictly smaller than
	$7$; i.e., that for any input $x=\langle v_1^1,v_1^2\rangle$, it holds
	that $N(x)=v_4^1 < 7$. This is encoded as a verification query by
	choosing a precondition that does not restrict the input, i.e.,
	$P=(true)$, and by setting $Q=(v_4^1\geq 7)$, which is the
	\textit{negation} of our desired property. For this verification
	query, a sound verifier will return \sat, alongside a feasible
	counterexample such as $x=\langle 0, 2\rangle$, which produces
	$v_4^1=22 \geq\ 7$. Hence, the property does not hold for this DNN.

	To date, the DNN verification community has focused primarily on 
	DNNs used for a single, non-reactive, invocation~\cite{ KaBaDiJuKo17, 
	GeMiDrTsChVe18, WaPeWhYaJa18,
		LyKoKoWoLiDa20, HuKwWaWu17}. Some work has been carried out on
	verifying DRL networks, which pose greater challenges: beyond the
	general scalability challenges of DNN verification, in DRL
	verification we must also take into account that agents typically
	interact with a reactive environment~\cite{CoMaFa21, BaGiPa21, ElKaKaSc21, 
	AmScKa21, JiTiZhWeZh22}.
	In particular, these agents are implemented with neural networks that are invoked multiple times, and the inputs of 
	each 
	invocation are usually
	affected by the outputs of the previous invocations. This fact
	aggregates the scalability limitations (because multiple invocations
	must be encoded in each query), and also makes the task of defining $P$ and 
	$Q$
	significantly more complex~\cite{AmScKa21}.

	\section{Case Study: Robotic Mapless Navigation}
	\label{sec:CaseStudy}

\begin{figure}[tb]
		\centering
		\includegraphics[width=0.9\linewidth]{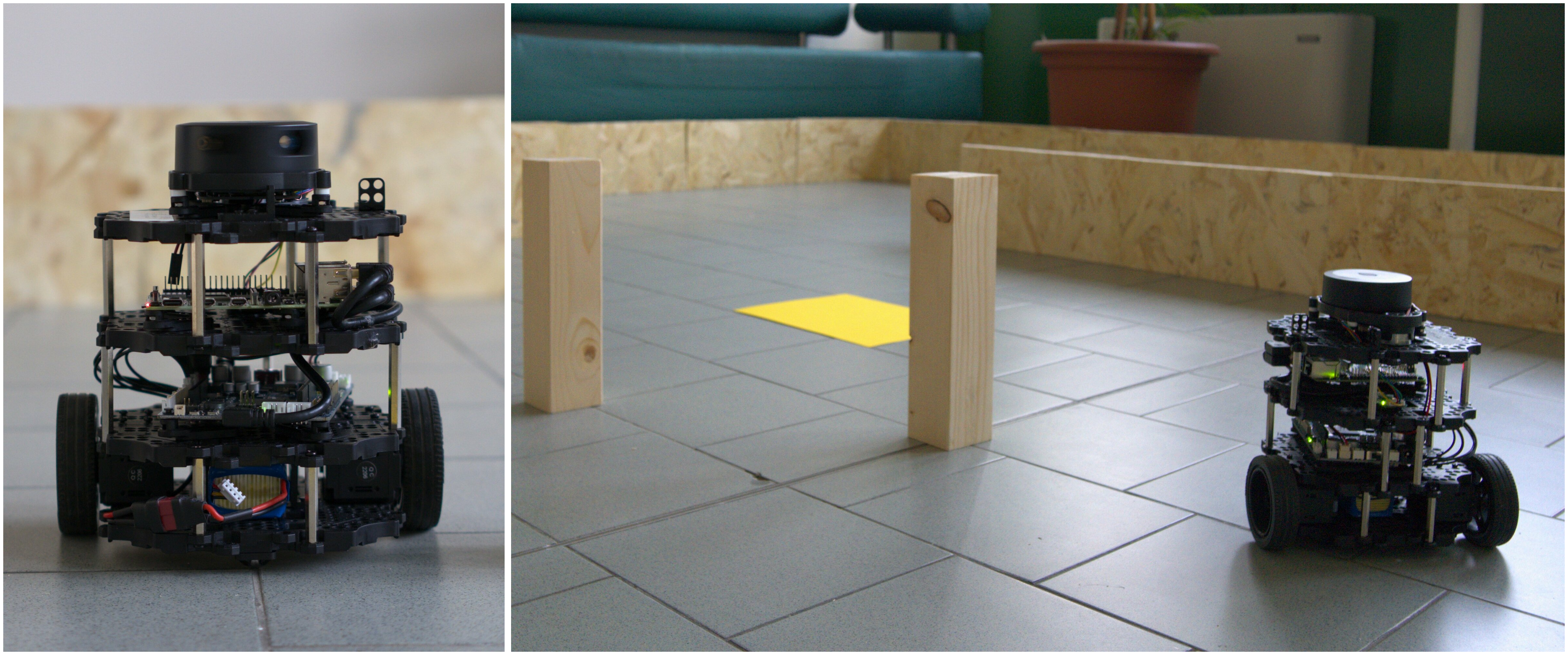}
		\caption{The \textit{Robotis Turtlebot 3} platform, navigating
			in an arena.  The image on the left depicts a static robot,
			and the image on the right depicts the robot moving towards
			the destination (the yellow square), while avoiding two
			wooden obstacles in its route.}
		\label{fig:environment:robot}
	\end{figure}

\begin{figure*}[t]
		\centering
		\includegraphics[width=1\textwidth]{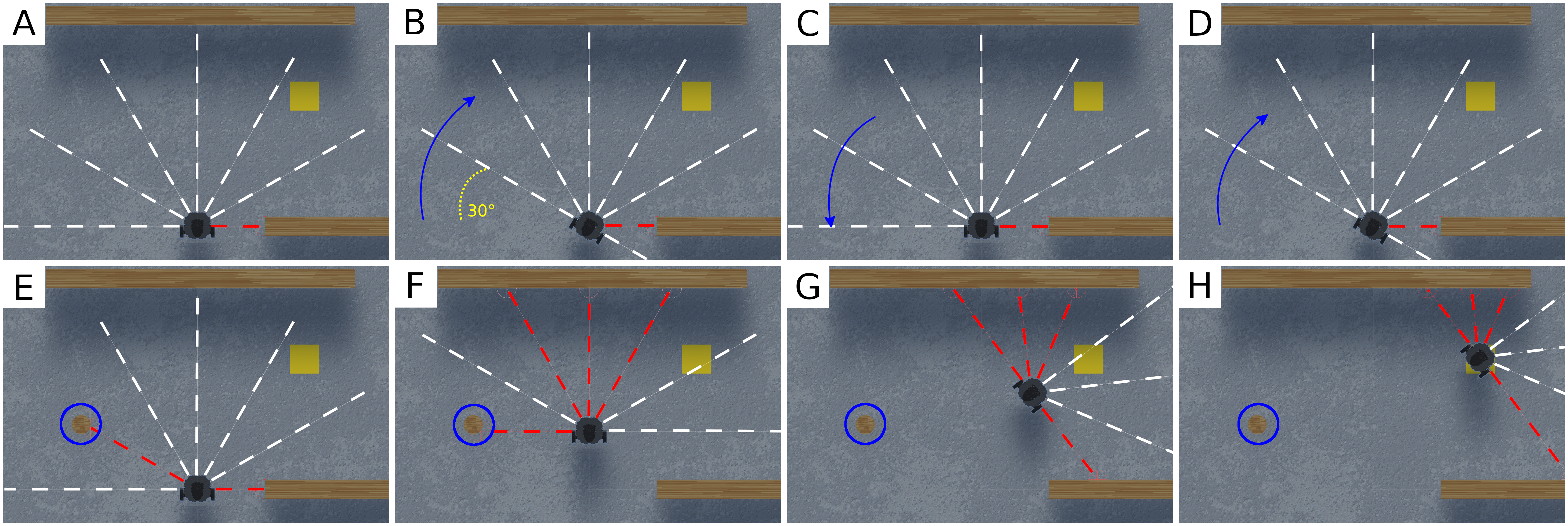}
		\caption{An example of a simulated Turtlebot entering a 2-step
			loop. The white and red dashed lines represent the lidar
			beams (white indicates ``clear'', and red indicates that an
			obstacle is detected).  The yellow square represents the
			target position; and the blue arrows indicate rotation.  In the 
			first 
			row, from left to right, the
			Turtlebot is stuck in an infinite loop, alternating between
			right and left turns.  Given the deterministic nature of
			the system, the agent will continue to select these same
			actions, ad infinitum.  In the second row, from left
			to right, we present an almost identical configuration, but
			with an obstacle located $30\degree$ to the robot's left (circled 
			in 
			blue). The presence of
			the obstacle changes the input to the DNN, and allows
			the Turtlebot to avoid entering the infinite loop; instead, it 
			successfully
			navigates to the target.  }
		\label{fig:properties:adversarial_trajectory}
	\end{figure*}

	\mysubsection{Robotis Turtlebot 3.}  In our case study, we focus on
	the \textit{Robotis Turtlebot 3} robot (\emph{Turtlebot}, for short),
	depicted in Fig.~\ref{fig:environment:robot}. 
	Given its relatively low cost and efficient sensor configuration, this 
	robot is 
	widely used in robotics research~\cite{NaShVa21, 
		AmSl19}.  In particular, this
	robotic platform has the actuators required for moving and turning, as
	well as multiple lidar sensors for detecting obstacles. These sensors
	use laser beams to approximate the distance to the nearest object in  
	their direction~\cite{YoTeOg14}. 
	In our experiments,
	we used a configuration with seven lidar sensors, each with a maximal
	range of one meter. Each pair of sensors are $30\degree$ apart, thus 
	allowing 
	coverage of $180\degree$.  The images in
	Fig.~\ref{fig:properties:adversarial_trajectory} depict a simulation
	of the Turtlebot navigating through an arena, and highlight the lidar
	beams. See Section~\ref{sec:appendix:RobotTechnicalSecifications} of
	the Appendix for additional details.
	
	\mysubsection{The Mapless Navigation Problem.}  \emph{Robotic
		navigation} is the task of navigating a robot (in our case, the
	Turtlebot) through an arena. The robot's goal is to reach a
	target destination while adhering to predefined restrictions;
	e.g., selecting as short a path as possible, avoiding obstacles, or
	optimizing energy consumption. In recent years, robotic navigation
	tasks have received a great deal of attention~\cite{WaToFi19, ZhSpJo17}, 
	primarily 
	due to their applicability to autonomous vehicles.
	
	We study here the popular \textit{mapless} variant of the robotic navigation
	problem, where the robot can rely only on local observations (i.e.,
	its sensors), without any information about the arena's structure or 
	additional data from external sources. 
	In this setting, which has been studied extensively~\cite{TaPaLi17}, the 
	robot 
	has access to the \emph{relative location} of the target, but does not 
	have a \emph{complete map} of the arena.
	This makes mapless navigation a partially observable problem, and among the 
	most challenging tasks to solve in the robotics domain~\cite{TaPaLi17, 
	ChFaFiFr19, ZhZhTaLiBu18, MaCoMaFa22}.

\begin{figure}[tb]
	\centering

	\subfloat[The DRL controller]{
		\hspace{-4mm}
		\includegraphics[width=0.49\textwidth]{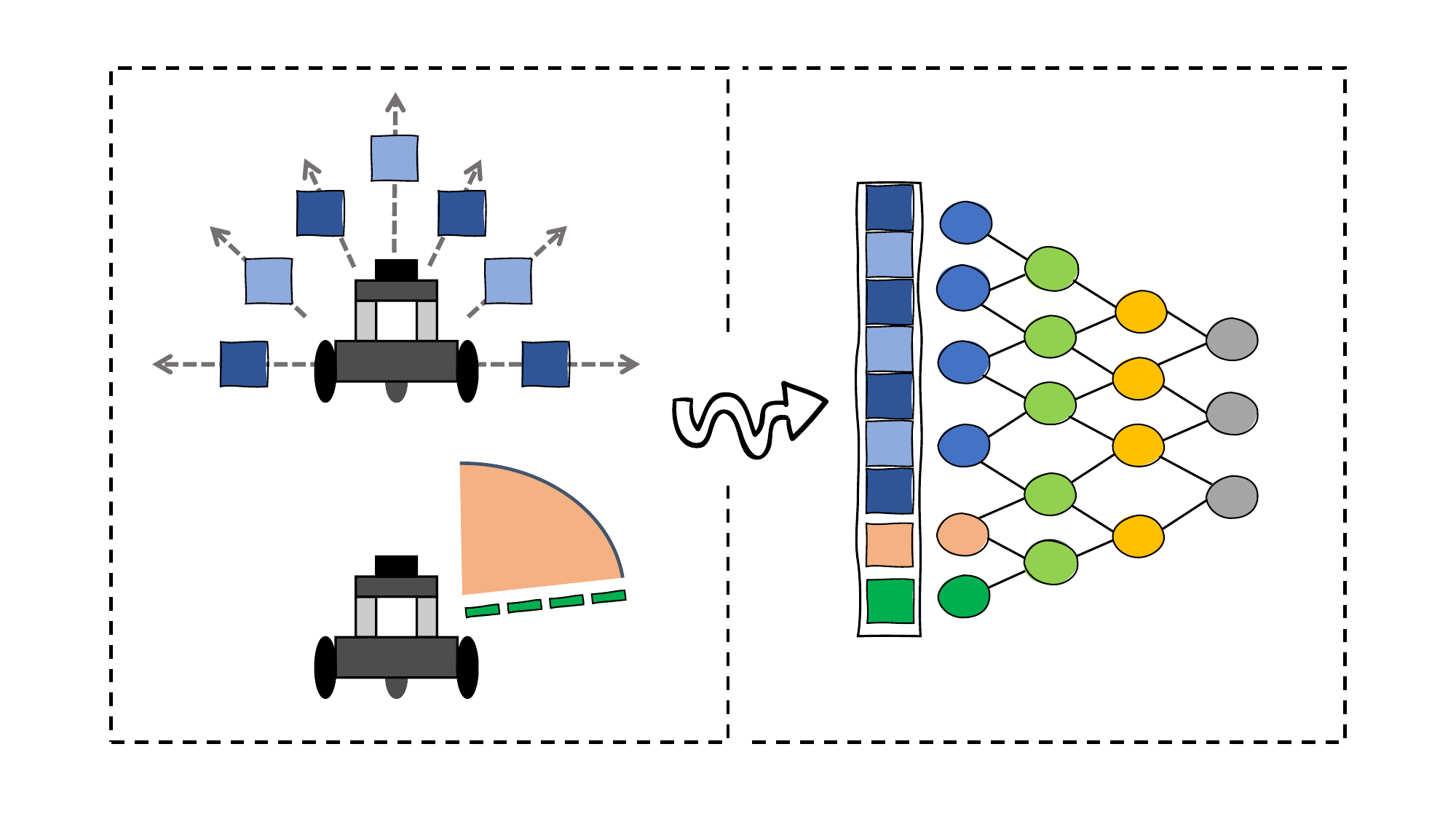}
		\label{fig:environment:RobotDRLStructure}
	}
	\subfloat[Average success rates]{
		\includegraphics[width=0.48\textwidth]{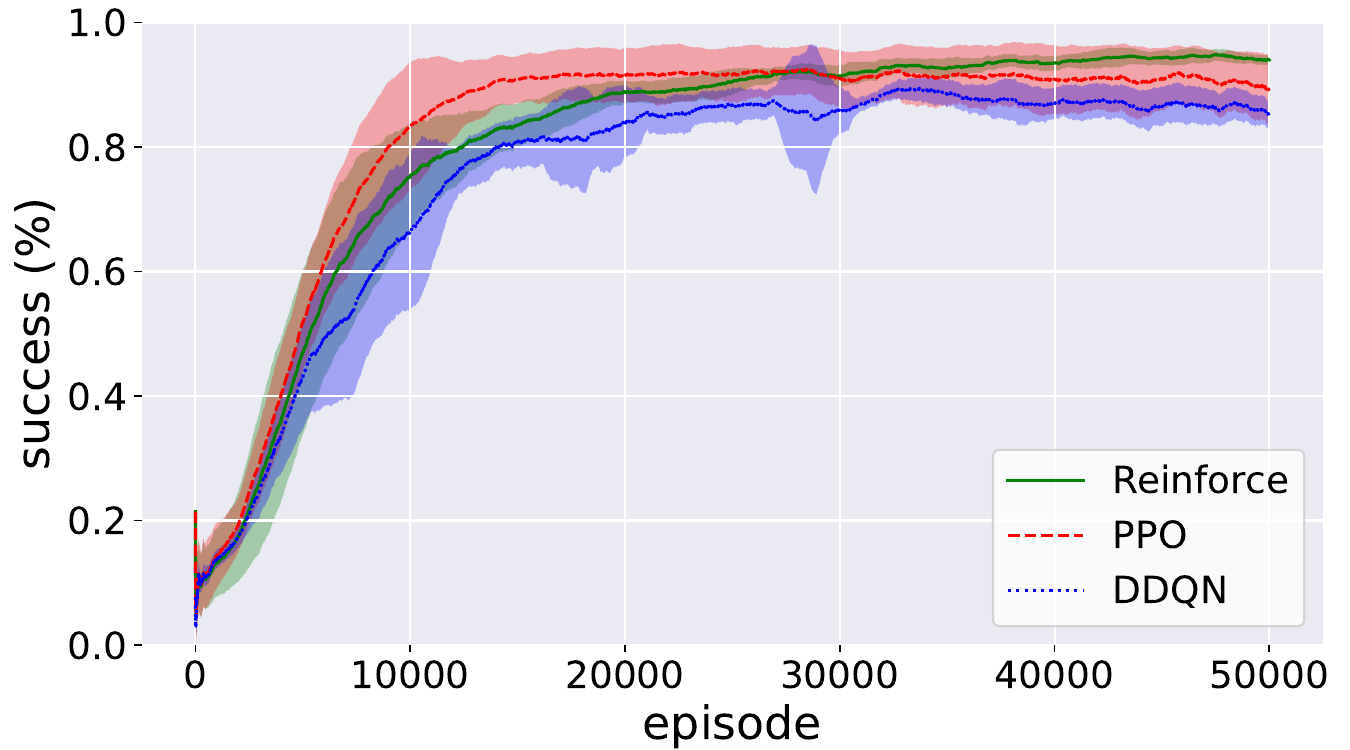}
		\label{fig:training:results}
	}
	\caption{
		(a) The DRL controller used for the robot in our case study. 
		The DRL has nine input neurons: seven lidar sensor readings (blue), 
		one input indicating the relative angle (orange) between the robot and 
		the target, and one input indicating the distance (green) between the 
		robot and the target.
		(b) The average success rates of models trained by each of the three 
		DRL training algorithms, per training episode. 
	}
\end{figure}

	
	\mysubsection{DRL-Controlled Mapless Navigation.}
	State-of-the-art solutions to mapless navigation suggest training a
	DRL policy to control the robot. Such DRL-based solutions have
	obtained outstanding results from a performance point of
	view~\cite{PfShTu18}. 
	For example, recent work by Marchesini et al.~\cite{MaFa20}
        has demonstrated how DRL-based agents can be
        applied to control the Turtlebot in a mapless navigation setting, by 
        training a 
        DNN with a
        simple architecture, including two hidden
        layers. Following this recent work, in
        our case study we used the following topology for DRL policies:

	\begin{itemize}
		\item An input layer with nine neurons. These include seven
		neurons representing the Turtlebot's lidar readings. The additional, 
		non-lidar inputs
		include one neuron representing the relative angle between the
		robot and the target, and one neuron representing the robot's
		distance from the target.  A scheme of the inputs appears in
		Fig.~\ref{fig:environment:RobotDRLStructure}.
		
		
		\item Two subsequent fully-connected layers, each consisting of $16$
                  neurons, and followed by a \relu{} activation layer.
		
		\item An output layer with three neurons, each corresponding to
		a different (discrete) action that the agent can choose to execute in
		the following step: move \forwardOutput{}, turn \leftOutput{}, or
		turn \rightOutput{}.\footnote{It has been shown that discrete
			controllers achieve excellent performance in robotic navigation,
			often outperforming continuous controllers in a large variety
			of tasks~\cite{MaFa20}.}
			
	\end{itemize}




	While the aforementioned DRL topology has been shown to be efficient for
	robotic navigation tasks, finding the optimal training algorithm
	and reward function is still an open problem.  As part of our work, we
	trained multiple \emph{deterministic} policies using the DRL algorithms 
	presented in
	Section~\ref{sec:Background}: DDQN~\cite{vHaGuSi16}, 
	Reinforce~\cite{ZhJoBr20}, 
	and PPO~\cite{ShWoDh17}.  For the reward
	function, we used the following formulation:
	\[
	\mathbb{R}_t = (d_{t-1} - d_{t}) \cdot \alpha - \beta
	,\] 
	where $d_t$ is the distance from the target at time-step $t$; $\alpha$ is a
	normalization factor used to guarantee the stability of the gradient; 
	and $\beta$ is a fixed value, decreased at each time-step, and resulting 
	in a total penalty proportional to the length of the path (by minimizing 
	this penalty, the agent is encouraged to reach the target quickly). In our 
	evaluation, we empirically selected $\alpha=3$ and $\beta=0.001$.
	Additionally, we added a final reward of $+1$ when the robot reached
	the target, or $-1$ in case it collided with an obstacle.  For
	additional information regarding the parameters chosen for the training
	phase, see Section~\ref{sec:appendix:TrainingTheModels} of the
	Appendix.

        
	
	\mysubsection{DRL Training and Results.}  Using the training
	algorithms mentioned in Section~\ref{sec:Background}, we trained a
	collection of DRL agents to solve the Turtlebot mapless navigation
	problem. 
	We ran a stochastic training process, and thus obtained varied 
	agents; 
	of these, we only kept those that achieved a success rate
	of at least $96\%$ during training. A total of $780$ models were
	selected, consisting of $260$ models per each of the three training 
	algorithms.
	More specifically, for each algorithm, all $260$ models were generated 
	from 
	$52$ random seeds. Each seed gave rise to a family of $5$ models, where 
	the individual family members differ in the number of training
	episodes used for training them. Fig.~\ref{fig:training:results} shows the 
	trained models'
	average success rate, for each algorithm used. 
	We note that PPO was generally the fastest to achieve high accuracy. 
	However, all 
	three training algorithms successfully produced highly accurate agents.


	\section{Using Verification for Model Selection}
	\label{sec:PropertiesAndPolicySelection}


       
        
	All of our trained models achieved very high success rates, and so, at
	face value, there was no reason to favor one over the other. However,
	as we show next, a verification-based approach can expose multiple
	subtle differences between them. As our evaluation criteria, we define
	two properties of interest that are derived from the main goals of
	the robotic controller: (i) reaching the target; and (ii) avoiding
	collision with obstacles.  	
	Employing verification, we use these
	criteria to identify models that may fail to fulfill their goals,
	e.g., because they collide with various obstacles, are overly
	conservative, or may enter infinite loops without reaching the
	target.
	We now define the properties that we used, and
	the results of their verification are discussed in
	Section~\ref{sec:Experiments}.  Additional details regarding the
	precise encoding of our queries appear in
	Section~\ref{sec:appendix:PropertiesForQueryEncoding} of the Appendix.
	
\begin{figure}
	\begin{center}
		\includegraphics[width=0.60\textwidth]{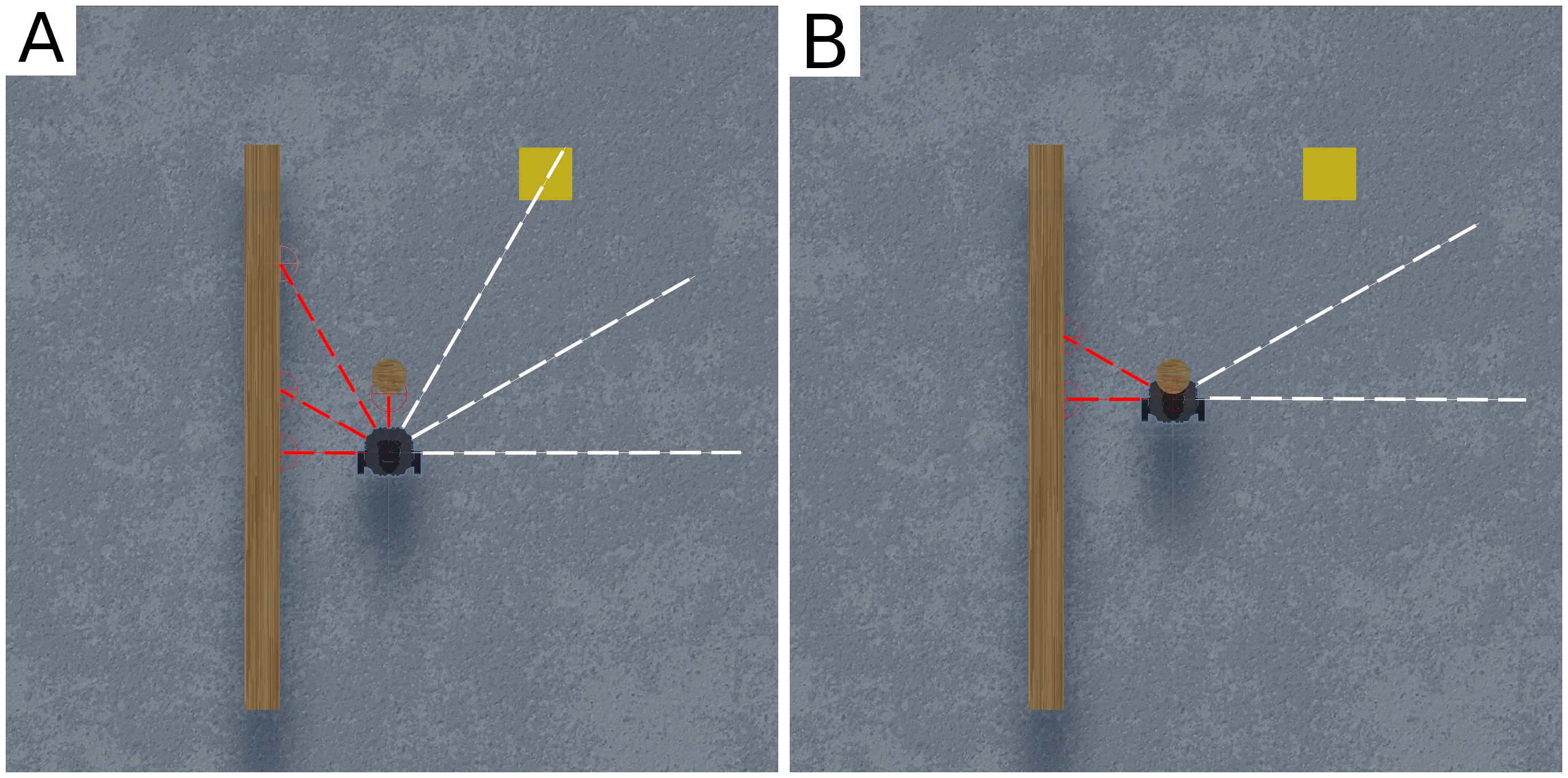}
	\end{center}
	\caption{Example of a single-step collision. The robot is not
		blocked on its right and can avoid the obstacle by turning
		(panel A),
		but it still chooses to move forward --- and collides (panel B).
	}
	\label{fig:environment:SimpleCollision}
    \end{figure}

	\mysubsection{Collision Avoidance.} Collision avoidance is a fundamental and
	ubiquitous safety property~\cite{ClHeVeBl18} for navigation agents. In the 
	context of
	Turtlebot, our goal is to check whether there exists a setting in which the
	robot is facing an obstacle, and chooses to move forward ---
	even though it has at least one other viable option, in the form of a
	direction in which it is not blocked.  In such situations, it is
	clearly preferable to choose to turn \leftOutput{} or \rightOutput{} instead
	of choosing to move \forwardOutput{} and collide.  See
	Fig.~\ref{fig:environment:SimpleCollision} for an illustration.
	
	Given that turning \leftOutput or \rightOutput produces an in-place 
	rotation 
	(i.e., the 
	robot does not change its position), the only action that can cause a 
	collision 
	is \forwardOutput. In particular, a collision can happen when an obstacle 
	is directly in front of the robot, or is slightly off to one side (just 
	outside 
	the front lidar's field of detection). More formally, we consider the safety
	property \emph{``the robot does not collide at the next step''}, with three 
	different 
	types of collisions:
	
	\begin{itemize}
		\item \forwardCollision{}: the robot detects an obstacle 
		straight ahead, but nevertheless makes a step forward and
		collides with the obstacle.
		
		\item \leftCollision{}: the robot detects an obstacle ahead
		and slightly shifted to the left (using the lidar beam that is 
		$30\degree$ to the left of the one pointing straight ahead),
		but makes a single step forward and collides with the
		obstacle. The shape of the robot is such that in this
		setting, a collision is unavoidable.
		
		\item \rightCollision{}: the robot detects an obstacle 
		ahead and slightly shifted to the right, but makes a single step 
		forward 
		and collides with the obstacle.
	\end{itemize}
	
	Recall that in mapless navigation, all observations are local --- the
	robot has no sense of the global map, and can encounter any possible
	obstacle configuration (i.e., any possible sensor reading). Thus, in
	encoding these properties, we considered a single invocation of the DRL
	agent's DNN, with the following constraints:
	
	\begin{enumerate}
		
		\item 
		All the sensors that are not in the direction of the obstacle receive a 
		lidar 
		input indicating that the robot can move either \leftOutput{} or 
		\rightOutput{} 
		without risk of collision. This is encoded by lower-bounding these 
		inputs.
		
		\item
		The single input in the direction of the obstacle is upper-bounded by a 
		value 
		matching the representation of an obstacle, close enough to the robot 
		so 
		that it will collide if it makes a move \forwardOutput.
		
		\item
		The input representing the distance to the target is lower-bounded,
		indicating that the target has not yet been reached (encouraging the
		agent to make a move). 
		
	\end{enumerate}
	
	The exact encoding of these properties is based on the physical
	characteristics of the robot and the lidar sensors, as explained in 
	Section~\ref{sec:appendix:RobotTechnicalSecifications} of the
	Appendix.

	\mysubsection{Infinite Loops.}  Whereas collision avoidance is the
	natural safety property to verify in mapless navigation controllers,
	checking that progress is eventually made towards the target is the
	natural liveness property. Unfortunately, this property is difficult
	to formulate due to the absence of a complete map. Instead, we settle
	for a weaker property, and focus on verifying that the robot does not
	enter infinite loops (which would prevent it from ever reaching the target).
	
	Unlike the case of collision avoidance, where a single step of the
	DRL agent could constitute a violation, here we need to reason about
	multiple consecutive invocations of the DRL controller, in order to
	identify infinite loops. This, again, is difficult to encode due to
	the absence of a global map, and so we focus on \emph{in-place} loops:
	infinite sequences of steps in which the robot turns \leftOutput and 
	\rightOutput,
	but without ever moving \forwardOutput, thus maintaining its current 
	location
	ad infinitum.
	
	Our queries for identifying in-place loops encode that: (i) the robot
	does not reach the target in the first step; (ii) in the following $k$
	steps, the robot never moves \forwardOutput, i.e., it only performs
	turns; and (iii) the robot returns to an already-visited
	configuration, guaranteeing that the same behavior will be
	repeated by our deterministic agents. The various queries differ in the 
	choice 
	of $k$, as well
	as in the sequence of turns performed by the robot. Specifically, we
	encode queries for identifying the following kinds of loops:
	\begin{itemize}
		\item \alternatingLoop{}: a loop where the robot performs an infinite
		sequence of
		$\langle \leftOutput, \rightOutput, \leftOutput, \rightOutput,
		\leftOutput...\rangle$ moves.  A query for identifying this loop
		encodes $k=2$ consecutive invocations of the DRL agent, after which
		the robot's sensors will again report the exact same reading,
		leading to an infinite loop. An example appears in
		Fig.~\ref{fig:properties:adversarial_trajectory}.  The encoding uses
		the ``sliding window'' principle, on which we elaborate later.
		
		\item \leftCycle{}, \rightCycle{}: loops in which the robot performs
		an infinite sequence of
		$\langle\leftOutput, \leftOutput, \leftOutput, \ldots\rangle$ or
		$\langle\rightOutput, \rightOutput, \rightOutput, \ldots\rangle$
		operations accordingly.
		Because the Turtlebot turns at a $30\degree$ angle, this loop is
		encoded as a sequence of $k=360\degree / 30\degree = 12$ consecutive 
		invocations of the DRL
		agent's DNN, all of which produce the same turning action (either 
		$\leftOutput$ or $\rightOutput$). Using the sliding window
		principle guarantees that the robot returns to the same exact
		configuration after performing this loop, indicating that it will
		never perform any other action.
	\end{itemize}	


	We also note that all the loop-identification queries include a condition
	for ensuring that the robot is not blocked from all directions. 
	Consequently,
	any loops that are discovered demonstrate a clearly suboptimal behavior.

	\mysubsection{Specific Behavior Profiles.}  
	In our experiments, we noticed that the safe policies, i.e., the ones that 
	do 
	not cause the 
	robot to collide, displayed a wide spectrum of different behaviors when 
	navigating to the target. 
	These differences occurred not only between policies that were trained by 
	different 
	algorithms, but also between policies trained by the same reward strategy 
	--- 
	indicating that these differences are, at least partially, due to the 
	stochastic 
	realization of the DRL training process.
	

	\begin{wrapfigure}{r}{0.5\textwidth}
		\vspace{-1.2cm}
		\begin{center}
			\includegraphics[width=0.48\textwidth]{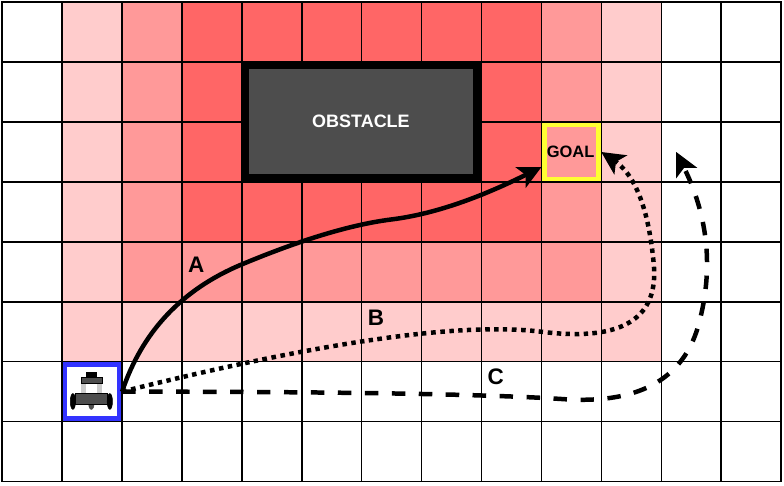}
		\end{center}
		\caption{
			Comparing paths selected by policies with different 
			\textit{bravery} levels. Path $A$ takes the Turtlebot close to the 
			obstacle (red area), and is the shortest. Path $B$ maintains a 
			greater distance from the obstacle (light red area), and is 
			consequently longer. Finally, path $C$ maintains such a significant 
			distance from the obstacle (white area) that it is unable to reach 
			the target.
		}
		\vspace{-6mm}
		\label{fig:verification:brave}
	\end{wrapfigure}
        Specifically, we noticed high variability in the length of the
	routes selected by the DRL policy in order to reach the given target:
	while some policies demonstrated short, efficient, paths that passed 
	very close to obstacles, other policies demonstrated a much more 
	conservative behavior, by selecting longer paths, and avoiding getting 
	close to obstacles (an example appears in 
	Fig.~\ref{fig:verification:brave}).
	
	Thus, we used our verification-driven approach to quantify how 
	conservative
	the learned DRL agent is in the mapless navigation setting. 
	Intuitively, a highly
	conservative policy will keep a significant safety margin from
	obstacles (possibly taking a longer route to reach its destination),
	whereas a 
	``braver'' and less conservative controller would risk venturing closer to
	obstacles.
	In the case of Turtlebot, 
	the preferable DRL policies are the ones that 
	guarantee the robot's safety (with respect to collision avoidance), and 
	demonstrate a high level of bravery
	--- as these policies tend to take 
	shorter, optimized paths (see path A in Fig.~\ref{fig:verification:brave}), 
	which 
	lead to reduced energy consumption over the entire trail. 
	
		
	Bravery assessment is performed by encoding verification queries that
	identify situations in which the Turtlebot \emph{can} move forward,
	but its control policy chooses not to. Specifically, we encode single
	invocations of the DRL model, in which we bound the lidar inputs to
	indicate that the Turtlebot is sufficiently distant from any obstacle
	and can safely move forward.  We then use the verifier to determine whether,
	in this setting, a \forwardOutput output is possible.  By altering and
	adjusting the bounds on the central lidar sensor, we can control how far
	away the robot perceives the obstacle to be. If we limit this distance
	to large values and the policy will still not move \forwardOutput{},
	it is considered conservative; otherwise, it is considered brave. By
	conducting a binary search over these bounds~\cite{AmScKa21}, we can
	identify the shortest distance from an obstacle for which the policy 
	\emph{safely} orders the robot
	to move \forwardOutput.  This value's inverse then serves as a bravery score 
	for that
	policy.

	\mysubsection{Design-for-Verification: Sliding Windows.}  A
	significant challenge that we faced in encoding our verification
	properties, especially those that pertain to multiple consecutive
	invocations of the DRL policy, had to do with the local nature of the
	sensor readings that serve as input to the DNN. Specifically, if
	the robot is in some initial configuration that leads to a sensor
	input $x$, and then chooses to move forward and reaches a successor
	configuration in which the sensor input is $x'$, some connection
	between $x$ and $x'$ must be expressed as part of the verification
	query (i.e., nearby obstacles that exist in $x$ cannot suddenly vanish in 
	$x'$). 
	In the absence of a global map, this is difficult to
	enforce.

	In order to circumvent this difficulty, we used the \emph{sliding
		window} principle, which has proven quite useful in similar
	settings~\cite{ElKaKaSc21, AmScKa21}. Intuitively, the idea is to
	focus on scenarios where the connections between $x$ and $x'$ are
	particularly straightforward to encode --- in fact, most of the sensor
	information that appeared in $x$ also appears in $x'$.  This approach
	allows us to encode multistep queries, and is also beneficial in
	terms of performance: typically, adding sliding-window constraints
	reduces the search space explored by the verifier, and expedites
	solving the query.
	
	In the Turtlebot setting, this is achieved by selecting a robot
	configuration in which the angle between two neighboring lidar sensors
	is identical to the turning angle of the robot (in our case,
	$30\degree$). This guarantees, for example, that if the central lidar
	sensor observes an obstacle at distance $d$ and the robot chooses to
	turn \rightOutput, then at the next step, the lidar sensor just to the left 
	of
	the central sensor must detect the same obstacle, at the same distance
	$d$.  More generally, if at time-step $t$ the 7 lidar readings (from
	left to right) are $\langle l_1,\ldots, l_7\rangle$ and the robot
	turns \rightOutput, then at time-step $t+1$ the 7 readings are
	$\langle l_2,l_3,\ldots,l_7,l_8\rangle$, where only $l_8$ is a new
	reading. The case for a \leftOutput turn is symmetrical. By
	placing these constraints on consecutive states encountered by the
	robot, we were able to encode complex properties that involve multiple
	time-steps, e.g., as in the aforementioned infinite loops. An illustration
	appears in Fig.~\ref{fig:properties:adversarial_trajectory}.

	\section{Experimental Evaluation}
	\label{sec:Experiments}
	
	Next, we ran verification queries with the aforementioned properties,
	in order to assess the quality of our trained DRL policies. The
	results are reported below. In many cases, we discovered 
	configurations in which the policies would cause the robot to collide
	or enter infinite loops; and we later validated the correctness of
	these results using a physical robot. We strongly encourage the reader
	to watch a short video clip that demonstrates some of these
	results~\cite{RobotYoutubeVideo}. Our code and
	benchmarks are also available online~\cite{ArtifactRepository}.	
    In our experiments, We used the \marabou verification 
    engine~\cite{KaHuIbHuLaLiShThWuZeDiKoBa19} as our backend, although other 
    engines could be used as well. For additional details regarding the 
    experiments, we refer the reader to 
    Section~\ref{sec:appendix:PropertiesForQueryEncoding} of the Appendix.

	\mysubsection{Model Selection.} In this set of experiments, we used
	verification to assess our trained models. Specifically, we used each of 
	the three training algorithms (DDQN, Reinforce, PPO) to train $260$ 
	models, creating a total of $780$ 
	models.
	For each of these, we verified
	six properties of interest: three collision properties
	(\forwardCollision, \leftCollision, \rightCollision), and three loop
	properties (\alternatingLoop, \leftCycle, \rightCycle), as described
	in Section~\ref{sec:PropertiesAndPolicySelection}. This gives a total
	of $4680$ verification queries.  We ran all queries with a \timeout
	value of $12$ hours and a \memout limit of $2G$; the results are
	summarized in Table~\ref{table:results:VerificationResults}. The
	single-step collision queries usually terminated within seconds,
	and the $2$-step queries encoding an \alternatingLoop usually
	terminated within minutes. The $12$-step cycle queries, which are
	more complex, usually ran for a few hours. $9.6\%$ of all queries hit
	the \timeout limit (all from the $12$-step cycle category), and none of
	the queries hit the \memout limit.\footnote{We note that two queries
		failed due to internal errors in \marabou.}

	\begin{table}[tb]
	\centering
	
	\begin{tabular}{l|cc|cc|cc}
		\hline
		\multicolumn{1}{c|}{} & 
		\multicolumn{2}{c|}{\texttt{\textbf{LEFT COLLISION}}} & 
		\multicolumn{2}{c|}{\texttt{\textbf{FORWARD COLLISION}}} & 
		\multicolumn{2}{c}{\texttt{\textbf{RIGHT COLLISION}}} \\ 
		\hline
		\texttt{Algorithm} & \sat & \unsat & \sat & \unsat & \sat & \unsat \\ 
		\hline
		DDQN & 259 & 1 & 248 & 12 & 258 & 2 \\
		Reinforce & 255 & 5 & 254 & 6 & 252 & 8\\
		PPO & 196 & 64 & 197 & 63 & 207 & 53 \\ 
		\hline
	\end{tabular}

	\vspace{3mm}
	
	\begin{tabular}{l|cc|cc|cc|c}
		\hline
		\multicolumn{1}{c|}{} & 
		\multicolumn{2}{c|}{\texttt{\textbf{ALTERNATING LOOP}}} & 
		\multicolumn{2}{c|}{\texttt{\textbf{LEFT CYCLE}}} & 
		\multicolumn{2}{c|}{\texttt{\textbf{RIGHT CYCLE}}} & 
		\texttt{INSTABILITY} \\  
		\hline
		\texttt{Algorithm} & \sat & \unsat & \sat & \unsat & \sat & \unsat & \# \texttt{alternations} \\ 
		\hline
		DDQN & 260 & 0 & 56 & 77 & 56 & 61 & 21 \\
		Reinforce & 145 & 115 & 5 & 185 & 120 & 97  & 10 \\
		PPO & 214 & 45 & 26 & 198 & 30 & 198  & 1 \\ 
		\hline
	\end{tabular}
	\vspace{3mm}
	\caption{
		Results of the policy verification queries. We verified six 
		properties over each of the 
		$260$ models trained per algorithm; \sat indicates that the
		property was violated, whereas \unsat indicates that it
		held (to reduce clutter, we omit \timeout and \fail results).
		The rightmost column reports the stability values of the
		various training methods. For the full results 
		see~\cite{ArtifactRepository}.
	}
	\label{table:results:VerificationResults}
	\vspace{-6mm}
\end{table}
	
	Our results exposed various differences between the trained models.
	Specifically, of the $780$ models checked, $752$ (over $96\%$)
	violated at least one of the single-step collision properties.  These
	$752$ collision-prone models include \textit{all} $260$ DDQN-trained
	models, $256$ Reinforce models, and $236$ PPO models.  Furthermore,
	when we conducted a model filtering process based on all six
	properties (three collisions and three infinite loops),
	we discovered that $778$ models out of the total of $780$ (over
	$99.7\%$!)  violated at least one property. The only two models that
	passed our filtering process were trained by the PPO algorithm.
	
	Further analyzing the results, we observed that PPO models tended to
	be safer to use than those trained by other algorithms:
	they usually had the fewest violations per property.  However, there
	are cases in which PPO proved less successful.  For example, our
	results indicate that PPO-trained models are more prone to enter an
	\alternatingLoop than those trained by Reinforce. Specifically,
	$214$ ($82.3\%$) of the PPO models have entered this
	undesired state, compared to $145$ ($55.8\%$) of the Reinforce
	models. We also point out that, similarly to the case with collision
	properties, \emph{all} DDQN models violated this property.
	
	Finally, when considering $12$-step cycles (either \leftCycle or
	\rightCycle), $44.8\%$ of the DDQN models entered such
	cycles, compared to $30.7\%$ of the Reinforce models, and just $12.4\%$ of
	the PPO models. In computing these results, we computed the fraction
	of violations (\sat queries) out of the number of queries that did not
	time out or fail, and aggregated \sat results for both cycle
	directions.
	
	Interestingly, in some cases, we observed a bias toward violating
	a certain subcase of various properties. For example, in the case of 
	entering
	full cycles --- although $125$ (out of $520$) queries indicated that
	Reinforce-trained agents may enter a cycle
	in either direction, in $96\%$ of these violations, the agent entered a
	\rightCycle. This bias is not present in models trained by the other
	algorithms, where the violations are roughly evenly divided between
	cycles in both directions.
	

	We find that our results demonstrate that different
        ``black-box'' algorithms generalize very differently with
        respect to various properties. In our setting, PPO produces
        the safest models, while DDQN tends to produce models with a
        higher number of violations. We note that this does not
        necessarily indicate that PPO-trained models perform better,
        but rather that they are more robust to corner cases.  Using
        our filtering mechanism, it is possible to select the safest
        models among the available, seemingly equivalent candidates.

	Next, we used verification to compute the bravery score of the various
	models.  Using a binary search,
	we computed for each model the minimal distance a dead-ahead obstacle
	needs to have for the robot to \emph{safely} move forward. The search range
	was $[0.18, 1]$ meters, and the optimal values were computed up to a
	$0.01$ precision (see
	Section~\ref{sec:appendix:PropertiesForQueryEncoding} of the Appendix
	for additional details). Almost all binary searches terminated within 
	minutes, 
	and none hit the \timeout threshold.
	
	%
	By first filtering the models based on their safe behavior, and then
	by their bravery scores, we are able to find the few models that are
	both safe (do not collide), and not overly conservative.  These models
	tend to take efficient paths, and may come close to an obstacle, but
	without colliding with it. We also point out that
	over-conservativeness may significantly reduce the success rate in
	specific scenarios, such as cases in which the obstacle is close to
	the target.  Specifically, of the only two models that survived the
	first filtering stage, one is considerably more conservative than the
	other --- requiring the obstacle to be twice as distant as the other,
	braver, model requires it to be, before moving forward. 





            	
	\mysubsection{Algorithm Stability Analysis.}  As part of our
	experiments, we used our method to assess the three training
	algorithms --- DDQN, PPO, and Reinforce. Recall that we used each
	algorithm to train $52$ families of $5$ models each, in which the
	models from the same family are generated from the same random seed, 
	but with a
	different number of training iterations. While all models obtained a high
	success rate, we wanted to check how often it occurred that a model
	successfully learned to satisfy a desirable property after some
	training iterations, only to forget it after additional iterations.
	Specifically, we focused on the $12$-step full-cycle properties
	(\leftCycle{} and \rightCycle{}), and for each family of $5$ models
	checked whether some models satisfied the property while others did
	not.

	
	We define a family of models to be \textit{unstable} in the case 
	where a property holds 
	in 
	the family, but ceases to hold 
	for another model from the same family with a higher number of training 
	iterations.
	Intuitively, this means that the model ``forgot'' a desirable property as 
	training progressed. The \emph{instability value} of each algorithm
	type is defined to be the number 
	of unstable 5-member families. 
	
	
	Although all three algorithms produced highly accurate models, they 
	displayed
	significant differences in the stability of their produced policies, as can 
	be 
	seen in the rightmost column of
	Table~\ref{table:results:VerificationResults}.  Recall that we trained
	$52$ families of models using each algorithm, and then tested their
	stability with respect to two properties (corresponding to the two full
	cycle types).  
	Of these, the DDQN models display $21$ \textit{unstable}
	alternations --- more than twice the number of alterations demonstrated by 
	Reinforce models ($10$), and significantly higher than the number of 
	alternations
	observed among the PPO models ($1$).  
	
	These results shed light on the nature of these training
	algorithms --- indicating that DDQN is a significantly less
	stable training algorithm, compared to PPO and Reinforce. This is in
	line with previous observations in non-verification-related
	research~\cite{ShWoDh17}, and is not surprising, as the primary
	objective of PPO is to limit the
	changes the optimizer performs between consecutive training
	iterations.

	
	\mysubsection{Gradient-Based Methods.}  We also conducted a thorough 
	comparison between our verification-based approach and competing 
	gradient-based methods. 
	Although gradient-based attacks are extremely scalable, our results 
	(summarized in Table~\ref{table:results:GradientBased} of the Appendix)
	show that they may miss many of the violations found by our complete,
	verification-based procedure. 
	For example, when searching for collisions, our approach discovered a total 
	of $2126$ \sat results, while 
	the gradient-based method discovered only  $1421$ \sat results --- a $33\%$ 
	decrease (!). 
	In addition, given that gradient-based methods are unable to return \unsat, 
	they are also incapable of proving that a property always holds, and hence 
	cannot formally guarantee the safety of a policy in question.
	Thus, performing model selection based on gradient-based methods could lead 
	to skewed results. We refer the reader to 
	Section~\ref{sec:appendix:comparsionToGradientBasedMethods} of the 
	Appendix, in which we elaborate on gradient attacks and the experiments we 
	ran, demonstrating the advantages of our approach for model selection, when 
	compared to gradient-based methods.
	
\section{Related Work}
\label{sec:RelatedWork}

Due to the increasing popularity of DNNs, the formal methods community has put 
forward a plethora of tools and approaches for verifying DNN 
correctness~\cite{Eh17,
	KaBaDiJuKo17, KaBaDiJuKo21, KaHuIbHuLaLiShThWuZeDiKoBa19,
	HuKwWaWu17, KuKaGoJuBaKo18, GoKaPaBa18, SiGePuVe19, GeMiDrTsChVe18,
	TjXiTe17, LoMa17}. Recently, the verification 
      of systems involving multiple DNN invocations, as well as hybrid
      systems with DNN components, 
has been receiving significant attention~\cite{FuPl18, VaPeWaNiSiKh22,BaGiPa21, 
	KaBaKaSc19,
	AmScKa21, DuChSa19, DuJhSaTi18b, 
	SuKhSh19}. Our work here is another step toward applying DNN verification
techniques to additional, real-world systems and properties of interest.

In the robotics domain, multiple approaches exist for
increasing the reliability of learning-based systems~\cite{RaAcAm19, 
	ZhZhWu21, WaSu20};
however, these methods are mostly heuristic in 
nature~\cite{GaFe15, AcHeTa17, MaCoFa21b}. 
To date, existing techniques rely mostly on Lagrangian multipliers~\cite{StAcAb20, RoGiRo21, 
	LiDiLi20}, and do not provide formal safety guarantees; rather, they 
	optimize 
the training in an attempt to learn the required policies~\cite{BrGrHa21}.
Other, more formal approaches focus solely on the systems' input-output 
relations~\cite{MaCoFa21a, CoMaFa21}, without considering multiple invocations 
of the agent and its interactions with the environment.
Thus, existing methods are not able to provide rigorous guarantees 
regarding the correctness of multistep robotic systems, and do not take into 
account sequential decision making --- which renders them insufficient 
for detecting various safety and liveness violations.

Our approach is orthogonal and complementary to many existing safe DRL 
techniques. Reward reshaping and shielding techniques 
(e.g.,~\cite{AlBlEhKoNiTo18}) improve safety by altering the training loop, but 
typically afford no formal guarantees. Our approach can be used to complement 
them, by selecting the most suitable policy from a pool of candidates, 
post-training. Guard rules and runtime shields are beneficial for preventing 
undesirable behavior of a DNN agent, but are sometimes less suited for 
specifying the \emph{desired} actions it should take instead. In contrast, our 
approach allows selecting the optimal policy from a pool of candidates, without 
altering its decision-making.

	\section{Conclusion}
	\label{sec:Conclusion}
	
	
Through the case study described in this paper, we demonstrate
that current verification technology is applicable to
real-world systems. We show this by applying verification
techniques for improving the navigation of DRL-based robotic
systems. 
We demonstrate how off-the-shelf verification
engines can be used to conduct effective model selection, as
well as gain insights into the stability of state-of-the-art
training algorithms. As far as we are aware, ours is the first
work to demonstrate the use of formal verification techniques on 
multistep properties of actual, real-world robotic navigation 
platforms. 
We also believe the techniques developed here will allow the use of 
verification to improve additional multistep systems (autonomous 
vehicles, surgery-aiding robots, etc.), in which we can 
impose a transition function between subsequent steps. 
However, our approach is limited by DNN-verification technology, which 
we use as a black-box backend. As that technology becomes more 
scalable, so will our approach. Moving forward, we plan to generalize our work 
to richer environments --- such as cases where a memory-enhanced agent 
interacts with moving objects, or even with multiple agents in the same arena, 
as well as running additional experiments with deeper networks, and more 
complex DRL systems. In addition, we see probabilistic verification of 
stochastic policies as interesting future work.

	\medskip
	\noindent
	\textbf{Acknowledgements.}  The work of Amir, Yerushalmi and Katz was
	partially supported by the Israel Science Foundation (grant number
	683/18). The work 
	of Amir was supported by a scholarship from the Clore Israel Foundation. 
	The work of Corsi, Marzari, and Farinelli was partially supported by the 
	``Dipartimenti di Eccellenza 2018-2022'' project, funded by the Italian 
	Ministry of Education, Universities, and Research (MIUR).
	The work of Yerushalmi and Harel was partially supported by a research 
	grant from the Estate of Harry Levine, the Estate of Avraham Rothstein, 
	Brenda Gruss and Daniel Hirsch, the One8 Foundation, Rina Mayer, Maurice 
	Levy, and the Estate of Bernice Bernath, grant 3698/21 from the ISF-NSFC 
	(joint to the Israel Science Foundation and the National Science Foundation 
	of China), and a grant from the Minerva foundation. 
	We thank Idan Refaeli for his contribution to this project.

	{
		\bibliographystyle{abbrv}
		\bibliography{references}
	}

	\newpage
	\appendix
	\renewcommand{\thesection}{\Alph{section}}
	
	\section*{\huge Appendices}
	
	\section{Training the DRL Models}
	\label{sec:appendix:TrainingTheModels}

	In this Appendix, we elaborate on the hyperparameters used for training, 
	alongside various implementation details. 
	The code is based on the 
	\textit{BasicRL} 
	baselines\footnote{\url{https://github.com/d-corsi/BasicRL}}.
	Our full code for training, as well as our original models, can be found in 
	our 
	publicly-available artifact accompanying this 
	paper~\cite{ArtifactRepository}.
	
	\mysubsection{General Parameters}
	\begin{itemize}
		
		\item \textit{episode limit}: 100,000
		\item \textit{number of hidden layers}: 2
		\item \textit{size of hidden layers}: 16
		\item \textit{gamma ($\gamma$)}: 0.99
		
	\end{itemize}

	 To facilitate our experiments, we followed~\cite{MaFa20} --- and used the 
	 same network topology, i.e., two fully-connected \relu layers, but focused 
	 on the smallest layer size that still achieved state-of-the-art results: 
	 $16$ neurons, instead of $64$ in~\cite{MaFa20}, as these achieved similar 
	 accuracy and allowed us to expedite verification.
	
	An additional difference, between our setting and the one appearing 
	in~\cite{MaFa20}, is that our agents have three outputs, instead of five.  
	However, this is beneficial, as our agents achieve 
	similar rewards to those reported 
	in~\cite{MaFa20}, and are more straightforward to verify 
	(“design-for-verification”).

	\mysubsection{DDQN Parameters}
	
	For the DDQN experiments, we rely on the Double DQN implementation (DDQN) 
	with 
	a ``soft'' update performed at each step. The network is a standard 
	feed-forward DNN, 
	without dueling architecture. 
	\begin{itemize}
		
		\item \textit{memory limit}: 5,000
		\item \textit{epochs}: 40
		\item \textit{batch size}: 128
		\item \textit{$\epsilon$-decay}: 0.99995
		\item \textit{tau ($\tau$)}: 0.005
		
	\end{itemize}

	\mysubsection{Reinforce Parameters}
	
	Our implementation is based on a version of Reinforce which 
	directly 
	implements the policy 
	gradient theorem~\cite{SuMcSi99}. The strategy for the update rule is a 
	pure 
	\text{Monte Carlo} 
	approach, without temporal difference rollouts.
	\begin{itemize}
		
		\item \textit{memory limit}: None
		\item \textit{trajectory update frequency}: 20
		\item \textit{trajectory reduction strategy}: sum
		
	\end{itemize}

	\mysubsection{PPO Parameters}
	
	For the value function estimation and the critic update, we adopted a 
	1-step 
	temporal difference strategy (TD-1). The update rule follows the guidelines 
	of the 
	\textit{OpenAI's Spinning 
		Up} 	
		documentation.\footnote{\url{https://spinningup.openai.com/en/latest/}} 

	\begin{itemize}
		
		\item \textit{memory limit}: None
		\item \textit{trajectory update frequency}: 20
		\item \textit{trajectory reduction strategy}: sum
		\item \textit{critic batch size}: 128
		\item \textit{critic epochs}: 60
		\item \textit{critic network size}: same as actor
		\item \textit{PPO clip}: 0.2
		
	\end{itemize}
	
	\mysubsection{Random Seeds for Reproducibility} 
	
	We now present a complete list of our seeds, sorted by value:
	
	[$35$, $47$, $67$, $68$, $73$, $76$, $77$, $81$, $90$, $91$, $114$, $128$, 
	$158$, $165$, $174$, $176$, $180$, $196$, $201$, $215$, $234$, $239$, 
	$240$, 
	$267$, $286$, $296$, $298$, $299$, $303$, $308$, $318$, $319$, $321$, 
	$343$, 
	$352$, $379$, $381$, $393$, $399$, $418$, $425$, $444$, $457$, $491$, 
	$502$, 
	$512$, $518$, $528$, $530$, $535$, $540$, $549$]
	
	At each execution, we fed the same seeds to procedures from the 
	\textit{Random}, \textit{NumPy} and 
	\textit{TensorFlow} Python modules.

	\newpage

	\section{Technical Specifications of the Robot}
	\label{sec:appendix:RobotTechnicalSecifications}
	
	In this Appendix, we describe the details regarding the technical 
	specifications of the robot, its sensors, and our design choices. 
	We performed our experiments on the 
	\textit{Robotis 
		Turtlebot 3}, in the \textit{burger} 
	version.\footnote{\url{https://www.robotis.us/turtlebot-3/}} Turtlebot is a 
	small research and development platform ($138$\textit{mm} x 
	$178$\textit{mm} x $192$\textit{mm}), that includes a set of various 
	sensors 
	for 
	mapping and navigation:
	
	\begin{itemize}
		\item $360\degree$ lidar sensor (with a maximal distance of $3.5$ m)
		\item Raspberry Pi 3 to control the platform
		\item Gyroscope, Accelerometer, and Magnetometer sensors
		\item (optional) Raspberry Pi Camera for perception
	\end{itemize}
	
	The manufacturer provides all the libraries and scripts for full 
	compatibility 
	with the standard Robotic Operating System (ROS). The robot is designed to 
	receive continuous input (\textit{linear} and \textit{angular} velocity). 
	In 
	our settings, we discretized the action space: therefore a single step 
	\forwardOutput 
	corresponds to a linear translation of $5$ cm, while a 
	\leftOutput/\rightOutput turn 
	produces 
	a 
	rotation of $30\degree$ in the desired direction.
	
	\begin{figure*}[h]
		\centering
		\includegraphics[width=0.5\textwidth]{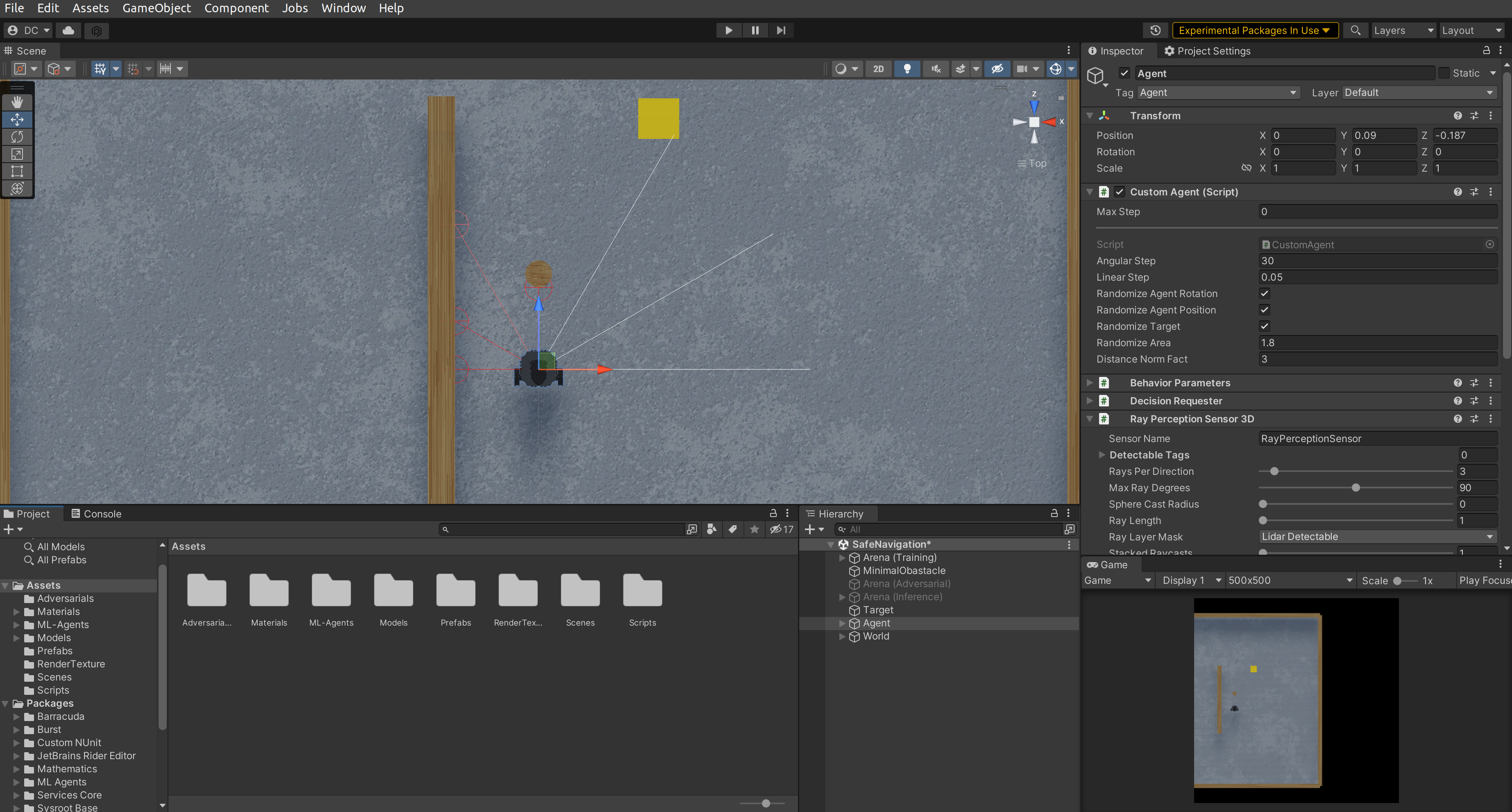}
		\caption{The \textit{Unity3D} engine with our simulation environment.}
		\label{fig:appendix:simulator}
	\end{figure*} 

	To perform the training of our robot, we rely on \textit{Unity3D}, a 
	popular engine originally designed for game development, that has 
	recently 
	been 
	adopted for robotics simulation~\cite{PoCoMa21, MaCoFa21a}. 
	In particular, the built-in physics engine, the powerful $3D$ rendering 
	algorithm 
	and the time control system (which allows speeding up the simulation by 
	more than 
	$10$ times), have made \textit{Unity3D} a very powerful tool in these 
	contexts~\cite{JuBeTe18}. 
	Fig.~\ref{fig:appendix:simulator} depicts an example of our Turtlebot3 
	environment in the \textit{Unity3D} simulator.
	A key advantage of mapless navigation is that the robot can access only 
	local observations. Thus, the robot can ``see'' only the distance to 
	the nearest collision point; hence the agent's decisions are agnostic 
	to the actual shape and size of the obstacles, and thus our system 
	is encouraged to 
	generalize to unseen environments during training.
	
	\newpage
	
	\section{Adversarial Trajectories}
	\label{sec:appendix:AdversarialTrajectories}
	
\begin{figure*}[h]
		\centering
		\includegraphics[width=0.9\textwidth]{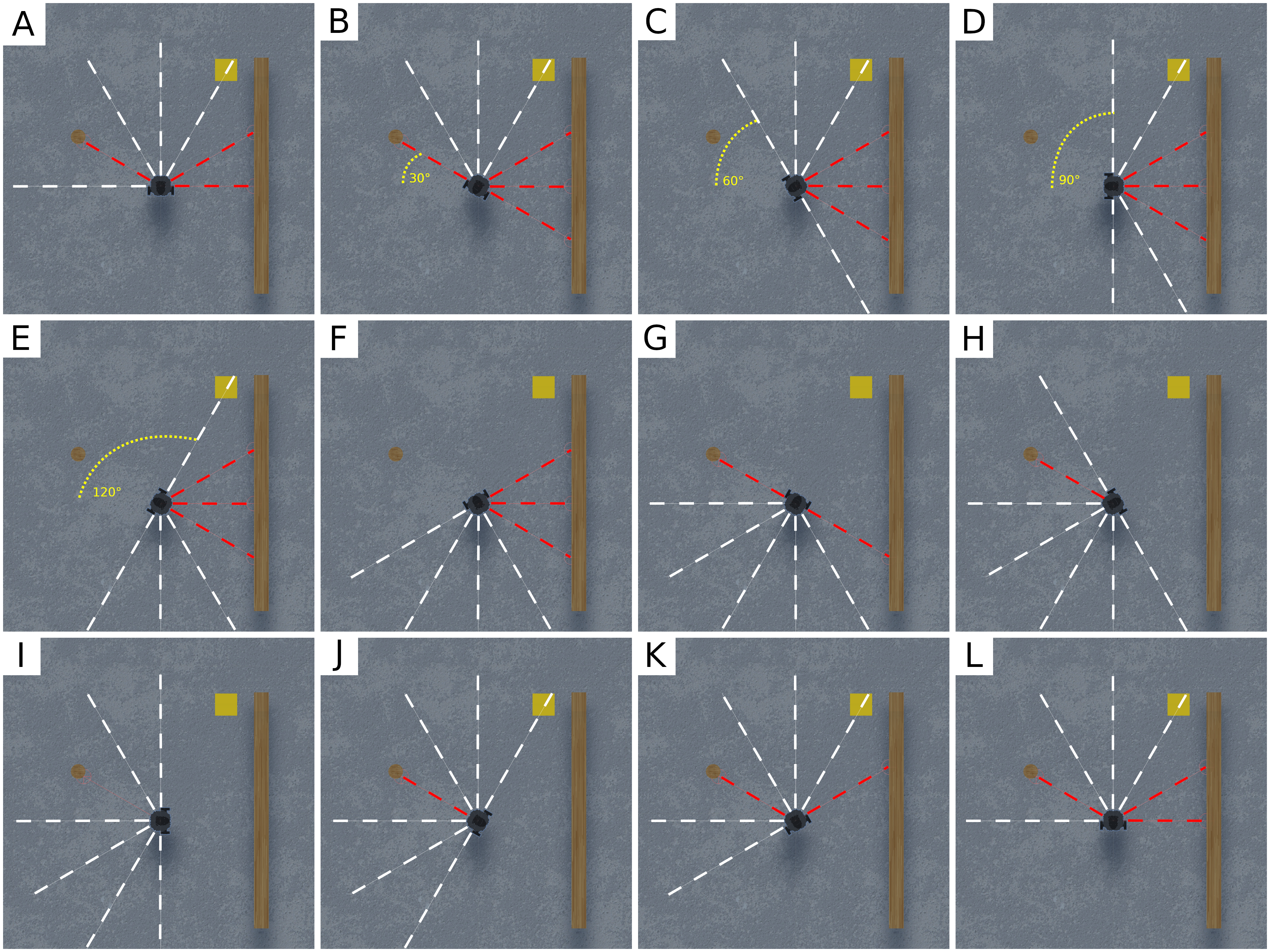}
		
		\caption{A 12-step trajectory of an infinite \rightCycle. The white 
			and red dashed lines represent the lidar scan (white indicates a 
			clear 
			path; red 
			indicates an obstacle is present), while the yellow square 
			represents 
			the 
			target position towards which the robot navigates. 
			The yellow dotted line represents the angular step the robot 
			performs 
			in the first 
			six steps. Given that the spacing between each lidar scan is 
			the 
			same as the 
			angular step size ($30\degree$), at each time-step it is possible 
			to 
			encode 
			the sliding 
			window for the state. The trajectory of a \leftCycle is symmetric.}
		
		\label{fig:appendix:RightCycleTrajectory}
	\end{figure*}

	In this Appendix, we depict a complete trajectory of a \rightCycle, as seen 
	in 
	Fig.~\ref{fig:appendix:RightCycleTrajectory}.
	We note that in this, and other, infinite-loop trajectories, it is crucial 
	to 
	encode that the first and last states of the trajectory are identical 
	(e.g., 
	Subfig. $A$ and Subfig. $L$ in 
	Fig.~\ref{fig:appendix:RightCycleTrajectory}). 
	Thus, given the deterministic nature of the 
	controller, the robot will repeat the same sequence of actions, 
	ending in an infinite loop, in which it constantly turns \rightOutput.

	\newpage
	
	\section{Encoding Verification Queries}
	\label{sec:appendix:PropertiesForQueryEncoding}
		
	In this Appendix, we formally define the complete list of properties we 
	analyzed in our experiments. All values are calibrated based on the 
	technical specifications of the robot, and the size of the discrete actions
	presented in this paper.
	
	Notice that, the symbol $x_n^t$ corresponds 
	to the input node $n$ of the network at time $t$, while $y_n^t$ is the 
	output node 
	$n$ at time $t$. Time ($t$) is omitted for the one-step properties (i.e., 
	$k$=$1$).

	\medskip
	\noindent
	\textbf{Determinsitic Policies and Stochastic Policies.}
	In order to compare DDQN to the stochastic policies (Reinforce and PPO), we 
	first trained the stochastic policies as usual, and then treated them as 
	deterministic. We note that in the case of stochastic policies, our 
	approach can only 
	detect whether a property \emph{may} be violated, and not how likely the 
	violation is.

	\medskip
	\noindent
	\textbf{Marabou.}
	All our queries were dispatched using the sound and complete \marabou 
	verification engine~\cite{KaHuIbHuLaLiShThWuZeDiKoBa19, 
		WuOzZeIrJuGoFoKaPaBa20}.   
	\marabou is a modern DNN verifier, whose core consists of a native
	Simplex solver, combined with abstraction and abstract-interpretation
	techniques~\cite{WaPeWhYaJa18, SiGePuVe19, ElGoKa20, OsBaKa22}, splitting
	heuristics~\cite{WuZeKaBa22}, optimization
	capabilities~\cite{StWuZeJuKaBaKo21},
	and enhanced to support varied
	activation functions~\cite{AmWuBaKa21}
	and recurrent networks~\cite{JaBaKa20}. \marabou has been previously
	applied to various verification-based tasks, such as network
	repair~\cite{ReKa22, GoAdKeKa20},
	network simplification~\cite{LaKa21, GoFeMaBaKa20}, ensemble 
	selection~\cite{AmZeKaSc22}, and more~\cite{AmFrKaMaRe23, 
		CaKoDaKoKaAmRe22, CoYeAmFaHaKa22, IsBaZhKa22, ZeWuBaKa22}. The \marabou 
		engine supports DNNs with ReLU layers, 
	max-pooling, convolution, absolute value, and sign layers; and also 
	supports sigmoids and softmax constraints. 
		

	\subsection{Property Constraints}
	
	\mysubsection{Trivial Bounds:} all the inputs of the network are normalized 
	in the 
	range $[0, 1]$. However, in the following cases, the actual bounds can be 
	tightened: (i) the true lower-bound for a lidar scan is $0.135$, given that 
	the 
	lidar scan is positioned in the center of the robot, therefore the size of 
	the 
	robot itself 
	constitutes a minimal distance from an obstacle; (ii) this lower bound can 
	be 
	further tightened to a value of $0.2$ to guarantee enough space for all the 
	possible 
	actions (i.e., \forwardOutput, \leftOutput, \rightOutput); (iii) for the 
	collision 
	properties, we found that $0.185$ is the minimal distance with which the 
	robot 
	can 
	make an action \forwardOutput, while avoiding a collision due to a 
	rotation; and (iv) to avoid configurations in which reaching the target 
	position
	is a trivial problem (e.g., a 1-step trajectory), we set a minimum distance 
	from 
	the target of $0.2$.
	
	\mysubsection{A complete list of constraints, per property} 
	
	Notice that all the 
	properties are encoded as a negation of the expected behavior 
	(see Sec.~\ref{sec:PropertiesAndPolicySelection} for details). Therefore a 
	\sat 
	assignment corresponds to a 
	violation of the required property.
	We note that the outputs $<y_0, y_1, y_2>$ correspond to the actions 
	$<\forwardOutput, \leftOutput, \rightOutput>$.
	
\begin{figure}
		\centering
		\includegraphics[width=0.2\textwidth]{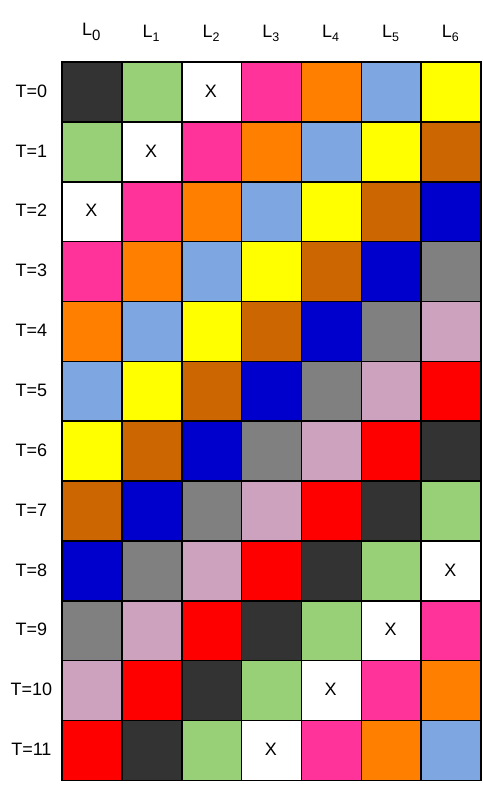}
		\caption{A visual scheme of the ``sliding window'' constraints for 
		encoding 
			a 12-step full cycle. In the case of encoding a \rightCycle, the 
			constraints 
			should be read from top to bottom. In the case of encoding a 
			\leftCycle, the 
			constraints should be read from bottom to top.}
		\label{fig:appendix:rainbow_loops}
              \end{figure}

	\begin{itemize}
		
		\item \textbf{\forwardCollision} ($k$=$1$): if the robot is closer than 
		$185$ mm 
		to an obstacle in front, never select the action \forwardOutput. If 
		violated, the robot collides in, at most, two steps.
		
		\begin{itemize}
			\item \textit{Precondition (P):} $x_i \in [0.2, 1] \text{ for } i = 
			[0, 
			1, 2, 
			4, 5, 6] \land x_3 \in [0.135, 0.185] \land x_7 \in [0, 1] \land 
			x_8 \in 
			[0.2, 
			1] $  
			
			\item \textit{Postcondition (Q):} $y_0 > y_1$ $\land$ $y_0 > y_2$
		\end{itemize}

		\item \textbf{\leftCollision} ($k$=$1$): if the robot is closer than 
		$185$ 
		mm 
		to an obstacle on the front-left, never select the action	
		\forwardOutput. 
		If violated, the robot collides in, at most, two steps.
		
		\begin{itemize}
			\item \textit{Precondition (P):} $x_i \in [0.2, 1] \text{ for } i = 
			[0, 
			1, 3, 
			4, 5, 6] \land x_2 \in [0.135, 0.185] \land x_7 \in [0, 1] \land 
			x_8 \in 
			[0.2, 
			1] $  
			
			\item \textit{Postcondition (Q):} $y_0 > y_1$ $\land$ $y_0 > y_2$
		\end{itemize}

		\item \textbf{\rightCollision} ($k$=$1$): if the robot is closer than 
		$185$ 
		mm 
		to an obstacle on the front-right, never select the action 
		\forwardOutput. If 
		violated, the robot will collide in, at most, two steps.
		
		\begin{itemize}
			\item \textit{Precondition (P):} $x_i \in [0.2, 1] \text{ for } i = 
			[0, 
			1, 2, 
			3, 5, 6] \land x_4 \in [0.135, 0.185] \land x_7 \in [0, 1] \land 
			x_8 \in 
			[0.2, 
			1] $  
			
			\item \textit{Postcondition (Q):} $y_0 > y_1$ $\land$ $y_0 > y_2$
		\end{itemize}

		\item \textbf{\alternatingLoop} ($k$=$2$): for all possible positions 
		of the
		target and the obstacles, never sequentially select the actions 
		$\rightOutput$ 
		and then immediately
		$\leftOutput$. If violated, the robot gets stuck in an infinite 2-step 
		loop.
		
		\begin{itemize}

			\item \textit{Precondition (P):} $x_i^{t} \in [0.2, 1] \text{ for } 
			i = 
			[0, 1, 2, 3, 4, 5, 6, 8] \land x_7^{t} \in [0, 1] \land 
			x_i^{t+1} \in [0.2, 1] \text{ for } i = [0, 1, 2, 3, 4, 5, 6, 8] 
			\land 
			x_7^{t+1} \in [0, 1] \land
			x_8^{t} = x_8^{t+1}$ $\land$ $x_i^{t}=x_{i+1}^{t}\text{ for } i = 
			[0, 1, 2, 3, 4, 
			5]$ 
			$\land$ $x_7^{t+1}=x_7^{t}\pm\frac{1}{12}$ 
			
			\item \textit{Postcondition (Q):} ($y_2^t > y_0^t$ $\land$ $y_2^t > 
			y_1^t$) $\land$ ($y_1^{t+1} > y_0^{t+1}$ $\land$ $y_1^{t+1} > 
			y_2^{t+1}$)
		\end{itemize}
		
		\noindent Given the symmetric nature of this property, the same 
		encoding 
		allowed us to check both the $\leftOutput$/$\rightOutput$ and 
		$\rightOutput$/$\leftOutput$ alternating loops. However, we note that 
		our 
		encodings 
		(arbitrarily) search for trajectories starting with a turn to the 
		\rightOutput.
		
		\item \textbf{\leftCycle} and \textbf{\rightCycle} ($k$=$12$): for all 
		possible positions of the
		target and the obstacles, if the robot has at least one escape 
		direction, never select in $12$ consecutive steps the action 
		$\leftOutput$ 
		(or $\rightOutput$). If this property is violated, the robot will get 
		stuck 
		in an infinite loop of $360\degree$ rotating counter-clockwise (or 
		clockwise).
		
		\begin{itemize}
			\item \textit{Precondition - bounds (P):} $x_i^{t} \in [0.2, 1] 
			\text{ 
				for } i = 
			[0, 1, 2, 3, 4, 5, 6, 8] \text{ and } t = [0, ..., 11] \land 
			x_7^{t} \in 
			[0, 1] \text{ for } t = [0, ..., 11] \land x_8^{t} = x_8^{t+1} 
			\text{ for 
			} t = [0, ..., 10]$
			\item \textit{Precondition - sliding window (P):} to better explain 
			the 
			property, we refer to Fig. \ref{fig:appendix:rainbow_loops} (for 
			\rightCycle the figure should be read from top to bottom; the 
			opposite 
			for 
			\leftCycle), which includes a scheme of the sliding window 
			encoding, 
			as 
			explained in Sec. 
			\ref{sec:PropertiesAndPolicySelection}. We note that sensors (in 
			different time-steps) are colored the same to represent that they 
			encode the same equality 
			constraint (e.g., $\langle L_1, T_0 \rangle = \langle L_0, T_1 
			\rangle$). 
			The input relative to the distance from the target (i.e., $x_8^t$) 
			must be 
			the same for all the $12$ steps, while the angle (i.e., $x_7^t$) 
			must 
			respect the property $x_7^{t+1}=x_7^{t}\pm\frac{1}{12}$ (depending 
			on 
			whether \rightCycle or \leftCycle is encoded).
			\item \textit{Postcondition (Q):} choose the same action 
			(\leftOutput 
			or 
			\rightOutput) for 12 consecutive steps.
		\end{itemize}
		
	\end{itemize}

\medskip
\noindent
\textbf{Note.}
The fact that the agent may enter infinite loops is not due to the turning 
angle matching the angle between consecutive lidar sensors; the same can happen 
when different angles are used, as in~\cite{MaFa20}. Configuring the lidar 
angles to match the turning angle is part of our methodology for facilitating 
verification (“sliding window”, for reducing the state space), but it does not 
cause the infinite loop, or any other property violation.

	\subsection{The Slack Margin}
	\label{sec:appendix:Winner_RunnerUp}
	
	When searching for adversarial inputs with formal verification 
	techniques, it 
	is common practice to search for \textit{strong} violations, i.e., 
	requiring 
	a 
	minimal difference between the original output and the output generated by 
	the 
	adversarial input. Formally, if the original 
	output is $y$, we consider a violation only if $y' > y + \gamma$, for some
	predefined margin (``slack'') $\gamma > 0$, and an output $y'$. 
	
	In our experiments, to conduct a fair comparison between models 
	trained by different algorithms (all outputting values in different ranges) 
	we 
	analyzed the empirical distribution of a random variable that consists of 
	the 
	difference between the ``winner'' (classified) output and the ``runner-up'' 
	(second highest) output.
	
\begin{figure*}[h]
		\centering
		\includegraphics[width=0.32\textwidth]{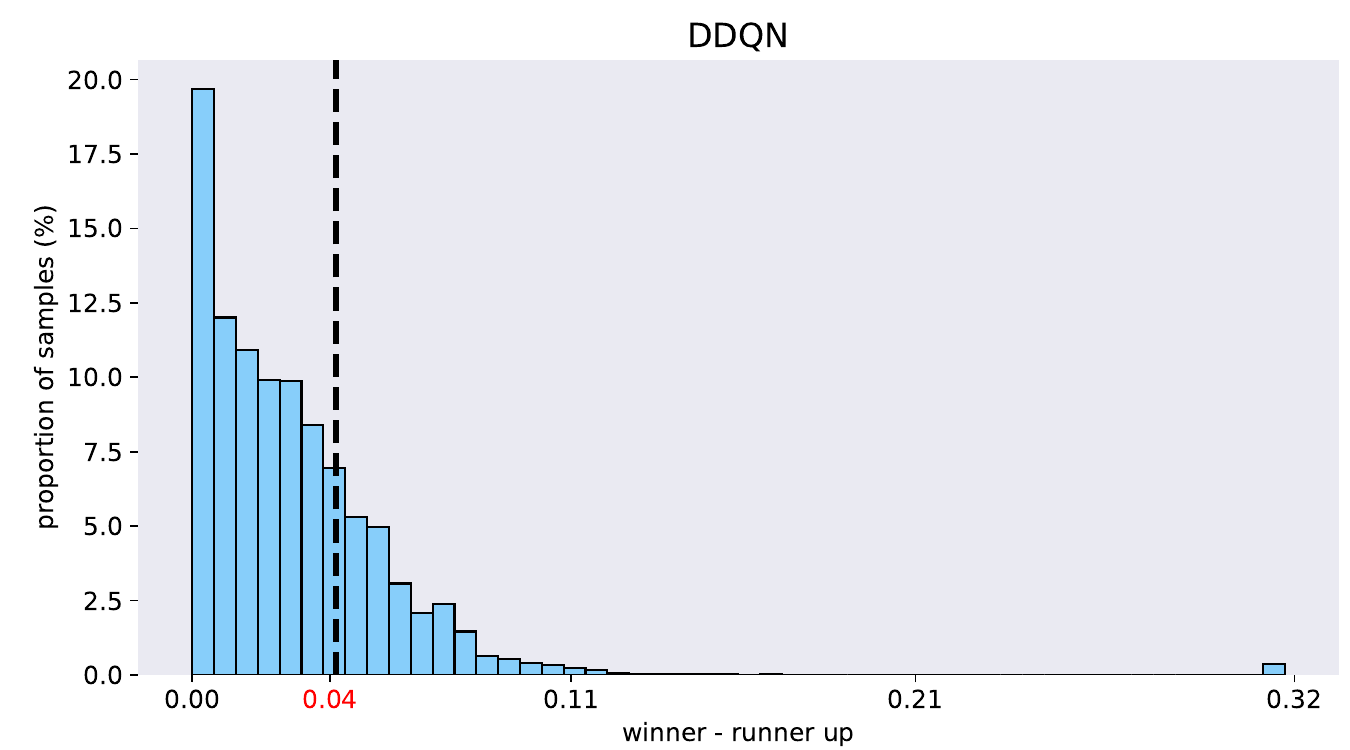}
		\includegraphics[width=0.32\textwidth]{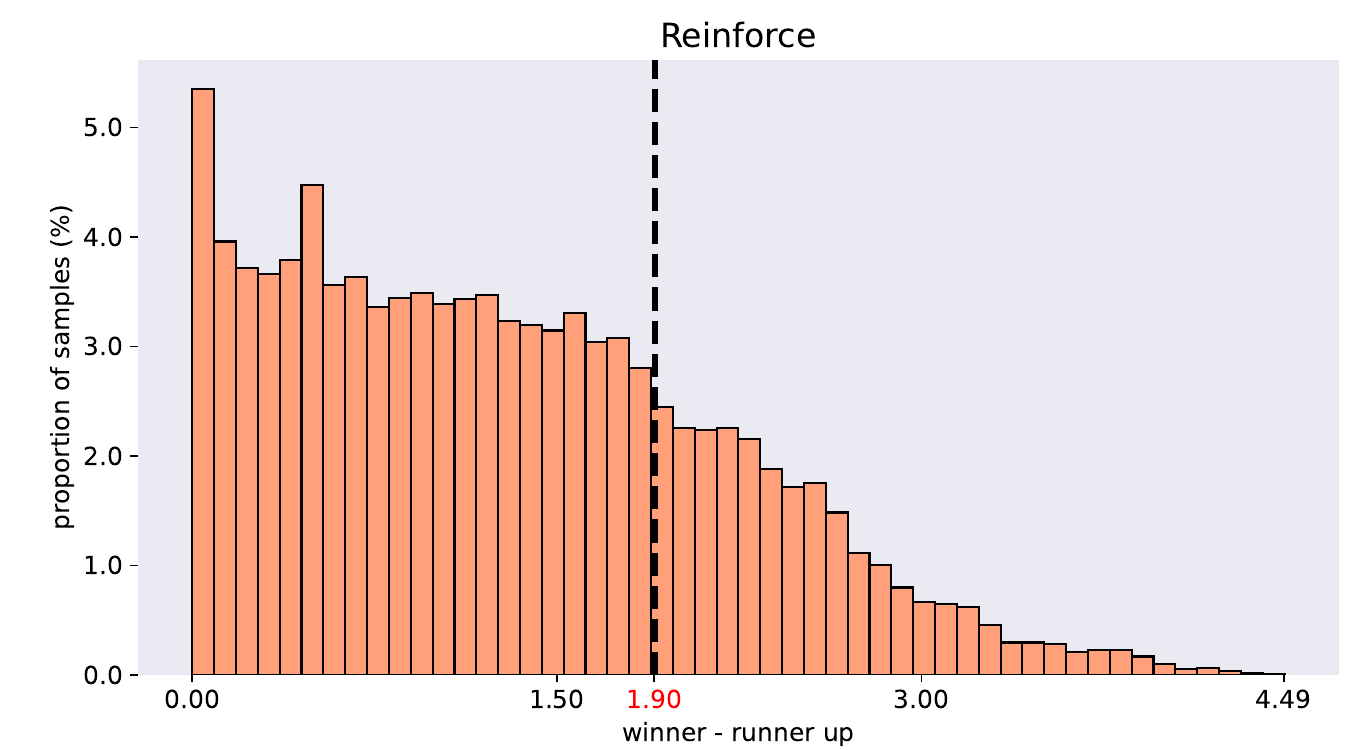}
		\includegraphics[width=0.32\textwidth]{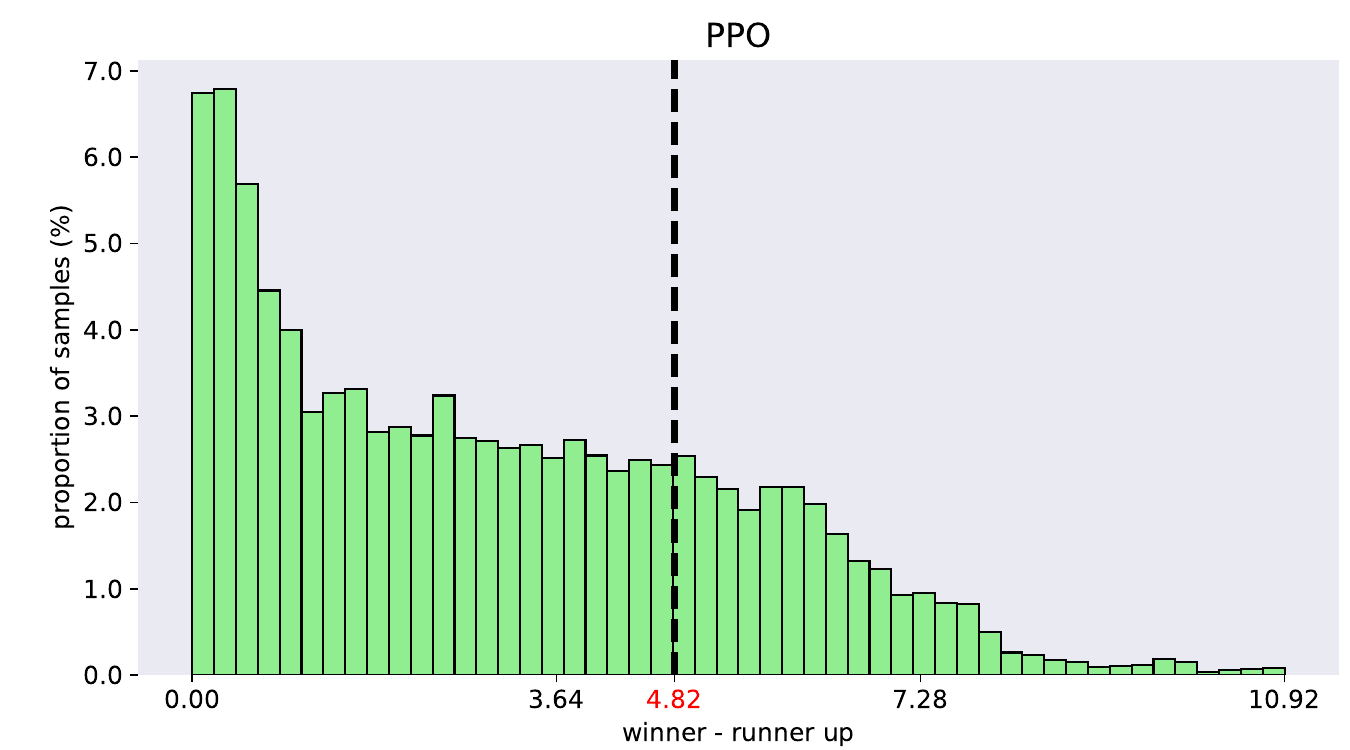}
		
		\caption{The empirical distribution of the difference between the 
		``winning'' (highest) 
			output and the second-highest output. 
			Each distribution is based on approximately $4000$ samples, taken 
			from 
			$N=100$ episodes for each algorithm.}
		\label{fig:appendix:distribution_plot}
	\end{figure*}

	As can be seen in Fig.~\ref{fig:appendix:distribution_plot}, the empirical 
	distribution for this random variable is different (both in the shape, and 
	values) per each of the training algorithms. 
	
	
	Given these results, to guarantee a fair comparison between the algorithms, 
	we 
	used a slack variable with a value corresponding to the \textit{75-th 
		percentile}, i.e., a value that is larger than $75\%$ of the samples. 
		The 
	values matching this percentile (per each training algorithm) are: 
	
	\begin{itemize}
		
		\item \textbf{DDQN}: $0.042$
		
		\item \textbf{Reinforce}: $1.904$
		
		\item \textbf{PPO}: $4.821$
		
		
	\end{itemize}

	
	These values were chosen to be the $\gamma$ slack values encoded for the 
	single-step ($k=1$) 
	verification queries, i.e., queries encoding: \forwardCollision, 
	\leftCollision 
	or \rightCollision.
	
	When encoding multistep properties (the three loop types) this 
	slack was divided by the number of $k$ steps encoded in the relevant query. 
	This is because each additional step adds constraints to the query 
	and so to balance the additional constraints --- the $\gamma$ slack is 
	reduced.

	For the original verification query encodings, we refer the reader to our 
	publicly-available artifact accompanying this 
	paper~\cite{ArtifactRepository}.

	\subsection{Bravery Property Results}
	\label{sec:appendix:Bravery}
		Our verification results indicate that:
		
		\begin{enumerate}
			\item the braver model may move forward when the distance is 
			slightly over $0.42$ meters; while
			
			\item  the over-conservative model never moves forward in
			cases where a similar obstacle is closer than $0.88$ meters.
		\end{enumerate}

	\newpage
	
	\section{Comparison to Gradient-Based Methods}
	\label{sec:appendix:comparsionToGradientBasedMethods}

	We also compared our
	verification-based method to state-of-the-art gradient attacks, which
	can also be used to search for property violations --- and hence, in
	model selection.  Gradient attacks are optimization methods, designed
	to produce inputs that are misclassified by
	the DNN~\cite{SzZaSuBrErGoFe13}. Intuitively, starting from some
	arbitrary input point, these methods seek perturbations
	that cause significant changes to the model's outputs --- increasing the
	value it assigns to outputs that correspond to some (targeted) incorrect
	label. These perturbations are discovered using a local search that
	performs gradient descent on a tailored loss function~\cite{GoShSz14}.
	It is common practice to consider a gradient-based search as
	successful if it finds a perturbed input on which the target label
	receives a score that is higher than the true label (by some predefined
	margin). Notice that, in contrast to complete verification
	methods, a gradient-based attack can only answer \sat or \timeout.
	
	In our evaluation, we ran the Basic Iterative
	Method (BIM)~\cite{KuGoBe18} ---  a popular gradient attack based
	on the Fast-Gradient Sign Method (FGSM)~\cite{GoShSz14, FeYi20}. 
	We ran the attack on all $780$ models, searching for violations of all 
	three 
	single-step safety properties. Each attack ran
	for $40$ iterations, with a step size of $0.01$. In 
	order to conduct a 
	fair comparison between this 
	gradient-based attack and our verification-based approach, we used a 
	heuristic of choosing the initial point for the attack to be the 
	\emph{center} of the bounds for each input. In addition, we used the same 
	slack margin by which the incorrect label should pass the correct one
	in all experiments, for both approaches. For additional details see 
	Section~\ref{sec:appendix:PropertiesForQueryEncoding} of the Appendix.
	
	\begin{table}[b]
		\centering
		\begin{tabular}{l|c|c|c}
			\hline
			\texttt{Algorithm} & 
			\texttt{Gradient} &  
			\texttt{Verification} &
			\texttt{Ratio (\%)} \\
			\hline
			DDQN & 405 & 765 & $53$ \\
			Reinforce & 671 & 761 & $88$ \\
			PPO & 345 & 600 & $58$ \\
			\hline
			\texttt{Total} & $1421$ & $2126$ & $67$ \\
		\end{tabular}
		\vspace{2mm}
		\caption{The number of counterexamples (\sat results)
			discovered using the gradient-based BIM
			method~\cite{KuGoBe18}, versus our verification-based approach.  
			The 
			rightmost
			column indicates the ratio of values in the two previous
			columns.}
		\label{table:results:GradientBased}
	\end{table}
	
	Although gradient-based methods are extremely scalable, they suffer from 
	many setbacks. Firstly, they tend to miss many violations otherwise found 
	by verification-driven approaches (up to a third of our violations were 
	missed, as seen in 	Table~\ref{table:results:GradientBased}). In addition, 
	gradient-based methods are incomplete, and thus, are unable to guarantee 
	that some properties always \emph{hold}, while sound and complete 
	verification tools may guarantee a property holds in various settings. 
	Another pitfall of gradient-based methods is their inadequacy to 
	check for adversarial inputs across multiple steps (loops, in our case), 
	bringing us to focus solely on collision properties in our comparison. 
	Hence, although gradient approaches are ill-suited for the task described, 
	we compared them to our approach, despite the inherent 
	limitations of such a comparison. Still, gradient attacks are the best tool 
	previously available, and so we believe this highlights the usefulness of 
	our approach.

\end{document}